\documentclass[10pt,twocolumn,letterpaper]{article}

\usepackage[pagenumbers]{cvpr} % To force page numbers, e.g. for an arXiv version

\definecolor{cvprblue}{rgb}{0.21,0.49,0.74}
\usepackage[pagebackref,breaklinks,colorlinks,allcolors=cvprblue]{hyperref}
\newcommand{\Tref}[1]{Table~\ref{#1}}

\newcommand{\fref}[1]{Fig.~\ref{#1}}

\usepackage{xcolor}
\newcounter{todos}
\AtEndDocument{\ifnum\value{todos}>0 \PackageWarning{TODOS}{There are \arabic{todos} todos left in this paper! Fix them before submitting the paper!} \fi}

\newcommand{\norm}[1]{\left\|#1\right\|}				% vector/matrix norm
\newcommand{\V}[1]{\ensuremath{\mathbf{#1}}}

\makeatletter
\DeclareRobustCommand\onedot{\futurelet\@let@token\@onedot}
\def\@onedot{\ifx\@let@token.\else.\null\fi\xspace}
\def\eg{\emph{e.g}\onedot} 
\def\ie{\emph{i.e}\onedot}

\makeatother
\usepackage[utf8]{inputenc} % allow utf-8 input
\usepackage[T1]{fontenc}    % use 8-bit T1 fonts
\usepackage{url}            % simple URL typesetting
\usepackage{booktabs}       % professional-quality tables
\usepackage{amsfonts}       % blackboard math symbols
\usepackage{nicefrac}       % compact symbols for 1/2, etc.
\usepackage{xcolor}         % colors
\usepackage{graphicx}
\usepackage{amsmath}
\usepackage{amssymb}
\usepackage{bm}
\usepackage{makecell}
\usepackage{multirow}
\usepackage{xspace}
\usepackage{colortbl}
\usepackage{wrapfig}
\usepackage{lipsum} % Just for generating dummy text, can be removed
\usepackage{threeparttable}
\usepackage{adjustbox}
\usepackage{makecell}
\usepackage{amsmath}
\usepackage{algorithm}
\usepackage{algpseudocode}

\definecolor{MyDarkRed}{rgb}{0.46, 0.16, 0.16}
\definecolor{MyDarkBlue}{rgb}{0.16, 0.16, 0.66}
\definecolor{MyPink}{rgb}{1.0, 0.702, 0.729} % #ffb3ba (255,179,186)
\definecolor{MyPeach}{rgb}{1.0, 0.875, 0.729} % #ffdfba (255,223,186)
\definecolor{MyLightYellow}{rgb}{1.0, 1.0, 0.729} % #ffffba (255,255,186)
\definecolor{MyLightGreen}{rgb}{0.729, 1.0, 0.788} % #baffc9 (186,255,201)
\definecolor{MyLightBlue}{rgb}{0.729, 0.882, 1.0} % #bae1ff (186,225,255)

\newcommand{\dmv}{DiLiGenT-MV~\cite{dmv}\xspace}

\newcommand{\neus}{NeuS~\cite{neus}\xspace}
\newcommand{\colmap}{COLMAP~\cite{colmap}\xspace}

\newcommand{\pandora}{PANDORA~\cite{pandora}\xspace}
\newcommand{\nersp}{NeRSP~\cite{nersp}}

\newcommand{\rnbneus}{RNb-Neus~\cite{rnbneus}\xspace}
\newcommand{\supernormal}{SuperNormal~\cite{supernormal}\xspace}

\newcommand{\nero}{NeRO~\cite{nero}\xspace}

\newcommand{\refneus}{Ref-NeuS~\cite{refneus}\xspace}
\newcommand{\pisr}{PISR~\cite{pisr}\xspace}
\newcommand{\sdm}{SDM-UniPS~\cite{sdm}\xspace}
\newcommand{\barf}{BARF~\cite{barf}\xspace}
\newcommand{\cfgs}{CF-3DGS~\cite{cfgs}\xspace}
\newcommand{\sparf}{SPARF~\cite{sparf}\xspace}
\newcommand{\nopenerf}{Nope-NeRF~\cite{nopenerf}\xspace}
\newcommand{\cogs}{COGS~\cite{cogs}\xspace}
\newcommand{\dust}{DUSt3R~\cite{dust3r}\xspace}
\newcommand{\stablenormal}{StableNormal~\cite{stablenormal}\xspace}
\newcommand{\rgreal}{RT3D\xspace}
\newcommand{\ours}{PMNI\xspace}
\newcommand{\pmni}{PMNI\xspace}

\DeclareMathOperator{\bce}{BCE}

\newcommand{\opacity}{\ensuremath{\alpha}\xspace}
\newcommand{\transmittance}{\ensuremath{T}\xspace}
\newcommand{\sharpness}{\ensuremath{s}\xspace}
\newcommand{\sigmoid}{\ensuremath{\phi_\sharpness}\xspace}

\newcommand{\pixel}{\V{p}\xspace}

\newcommand{\surface}{\ensuremath{\mathcal{M}}\xspace}

\newcommand{\point}{\ensuremath{\V{x}}\xspace}
\newcommand{\normal}{\ensuremath{\V{n}}\xspace}

\newcommand{\loss}{\mathcal{L}\xspace}

\newcommand\blfootnote[1]{%
	\begingroup
	\renewcommand\thefootnote{}\footnote{#1}%
	\addtocounter{footnote}{-1}%
	\endgroup
}

\def\paperID{6222} % *** Enter the Paper ID here
\def\confName{CVPR}
\def\confYear{2025}

\title{PMNI: Pose-free Multi-view Normal Integration for Reflective and Textureless Surface Reconstruction}

\author{Mingzhi Pei$^{1}$~~~~~~~~~Xu Cao$^{2}$~~~~~~~~~Xiangyi Wang$^{1}$~~~~~~~~~Heng Guo$^{1*}$~~~~~~~~~Zhanyu Ma$^{1}$\\
{$^1$Beijing University of Posts and Telecommunications}~~~~~~~
{$^2$Independent Researcher}\\
\small{\texttt{\{pmz, el777216wxy, guoheng, mazhanyu\}@bupt.edu.cn~~~~~~xucao.42@gmail.com}}
}

\begin{document}
\maketitle
\begin{abstract}
Reflective and textureless surfaces remain a challenge in multi-view 3D reconstruction.
Both camera pose calibration and shape reconstruction often fail due to insufficient or unreliable cross-view visual features.
To address these issues, we present PMNI (Pose-free Multi-view Normal Integration), a neural surface reconstruction method that incorporates rich geometric information by leveraging surface normal maps instead of RGB images. 
By enforcing geometric constraints from surface normals and multi-view shape consistency within a neural signed distance function (SDF) optimization framework, PMNI simultaneously recovers accurate camera poses and high-fidelity surface geometry. 
Experimental results on synthetic and real-world datasets show that our method achieves state-of-the-art performance in the reconstruction of reflective surfaces, even without reliable initial camera poses.
\end{abstract}    
\blfootnote{$^{*}$ Corresponding author.}
\blfootnote{Released at: \href{https://github.com/pmz-enterprise/PMNI}{https://github.com/pmz-enterprise/PMNI}}

\section{Introduction}
\label{sec:intro}

\newcommand{\imagesize}{0.25}
\begin{figure}
	\small
    \begin{tabular}{@{}c@{}@{}c@{}@{}c@{}@{}c@{}@{}c@{}}
         \includegraphics[width = 0.24\linewidth,   trim = 0pt 50pt 0pt 0pt, clip]{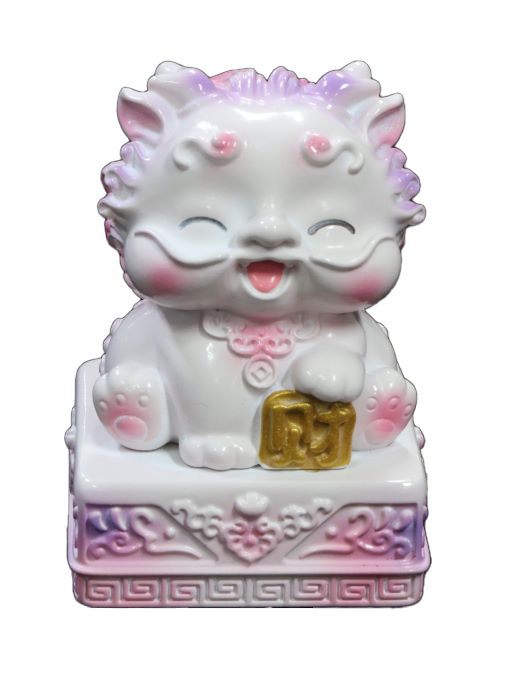}
         &\includegraphics[width = \imagesize\linewidth,  trim = 330pt 150pt 330pt 100pt, clip]{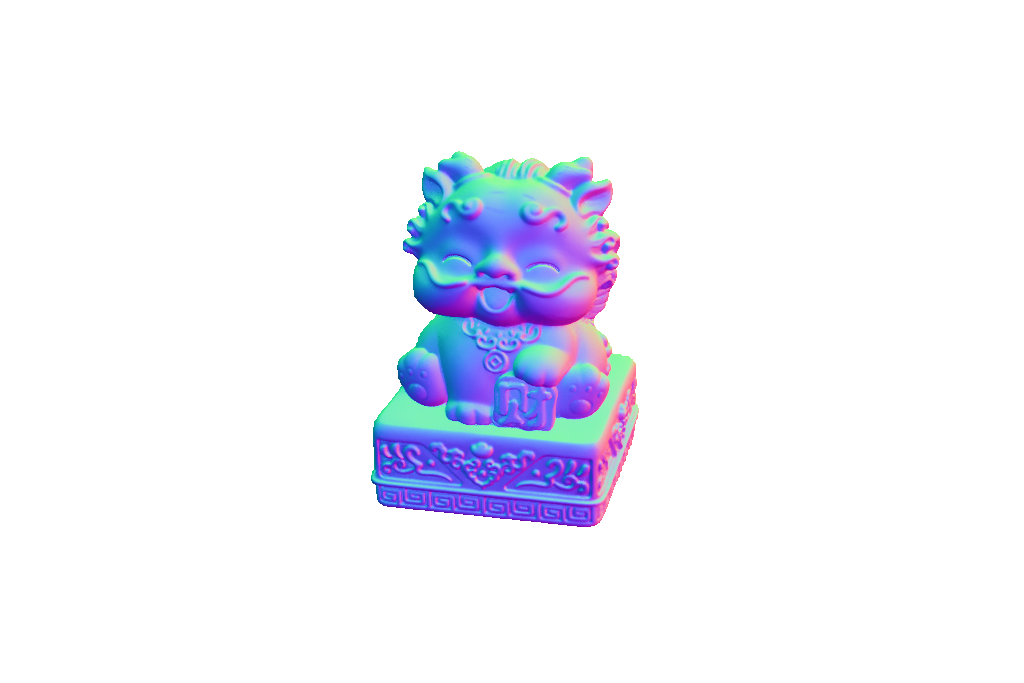}
         &\includegraphics[width = \imagesize\linewidth,  trim = 290pt 180pt 390pt 90pt, clip]{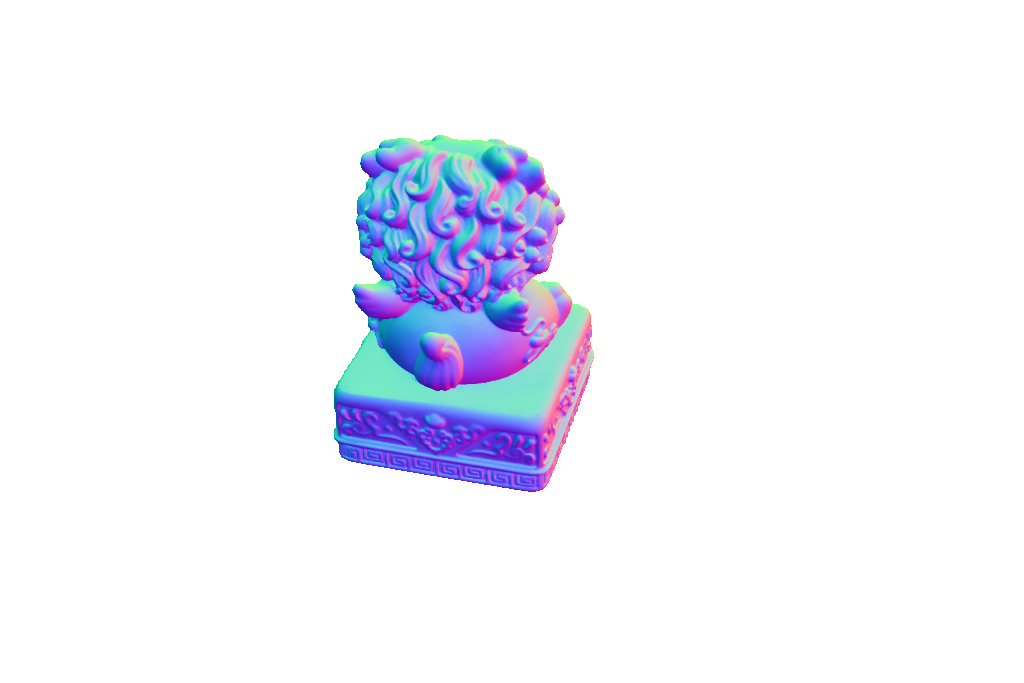}
         &\includegraphics[width = \imagesize\linewidth,  trim = 330pt 170pt 330pt 80pt, clip]{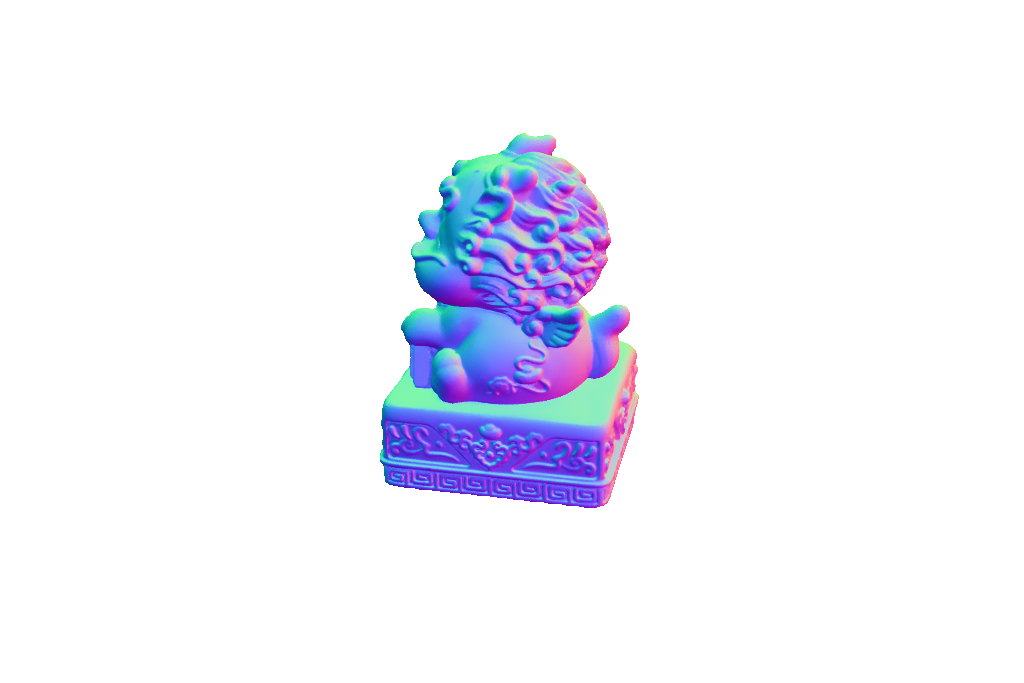}
          \\
          \midrule
         \includegraphics[width = \imagesize\linewidth,  trim = -20pt -20pt -20pt -20pt, clip]{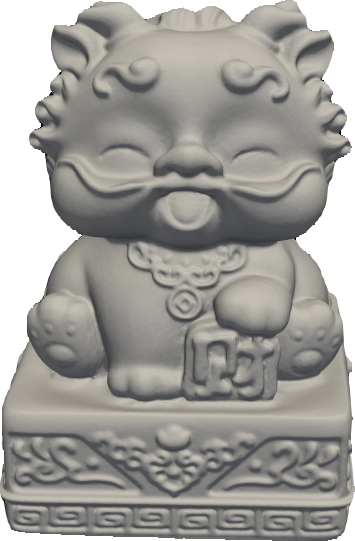}
         &\includegraphics[width = \imagesize\linewidth,  trim = -20pt -20pt -20pt -20pt, clip]{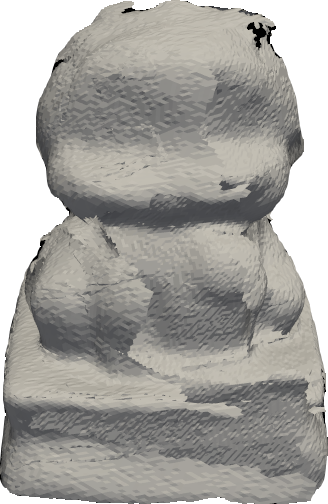}
         &\includegraphics[width = \imagesize\linewidth,  trim = -20pt -20pt -20pt -20pt, clip]{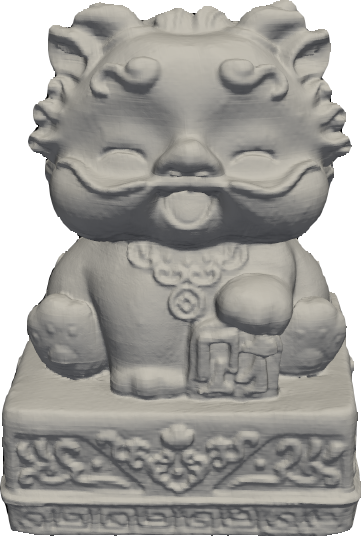}
         &\includegraphics[width = \imagesize\linewidth,  trim = -20pt -20pt -20pt -20pt, clip]{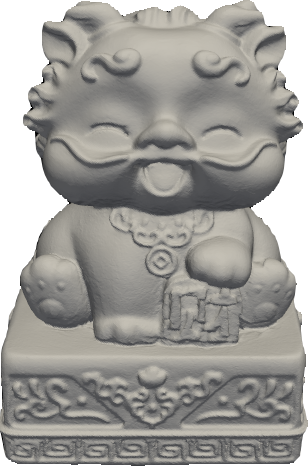}
          \\
           GT Mesh&  \dust & Ours & \supernormal 
          \\

         \includegraphics[width = \imagesize\linewidth,  trim = 0pt 0pt 0pt 20pt, clip]{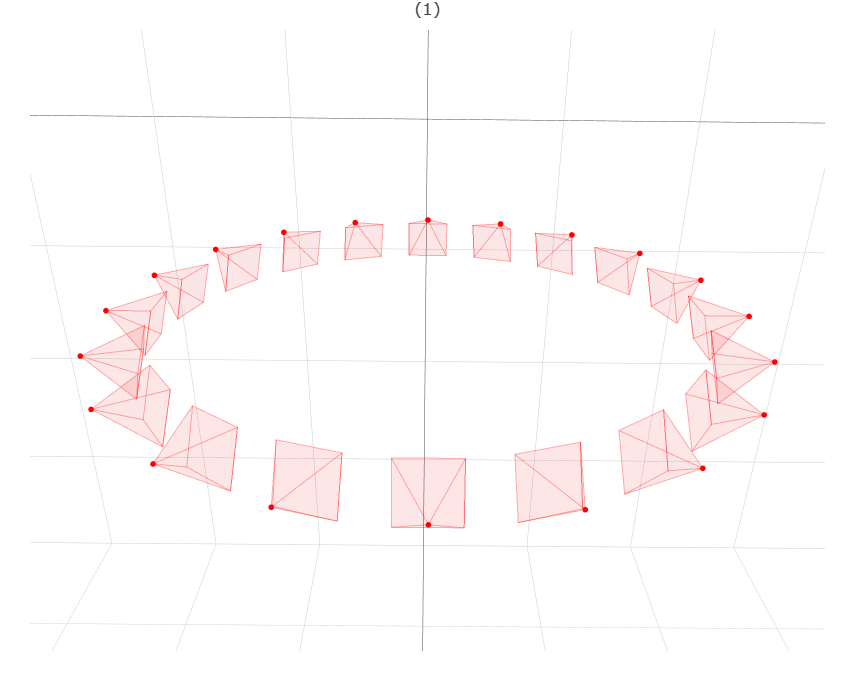}
      	&\includegraphics[width = \imagesize\linewidth,  trim = 0pt 0pt 0pt 20pt, clip]{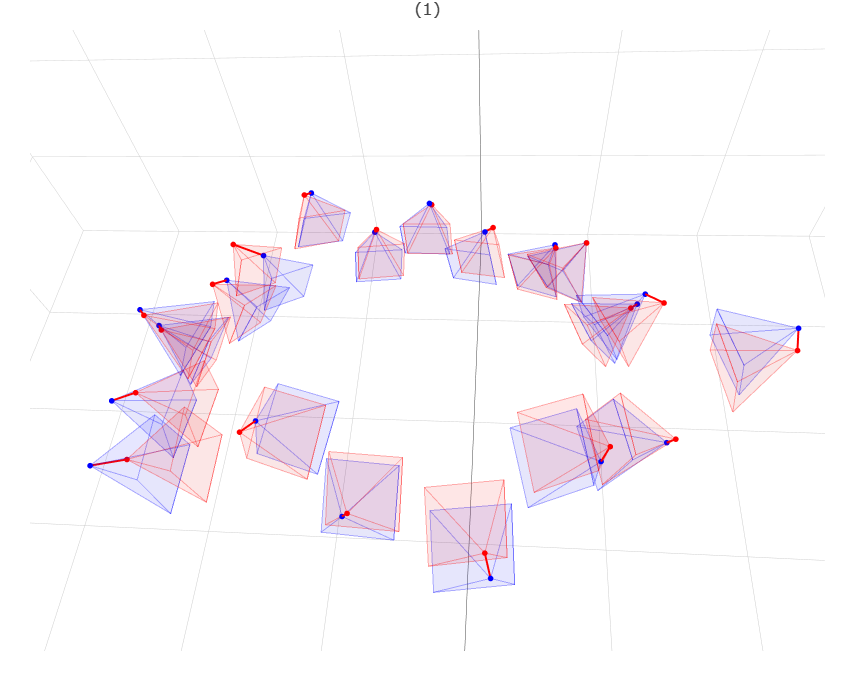}
      	&\includegraphics[width = \imagesize\linewidth,  trim = 0pt 0pt 0pt 20pt, clip]{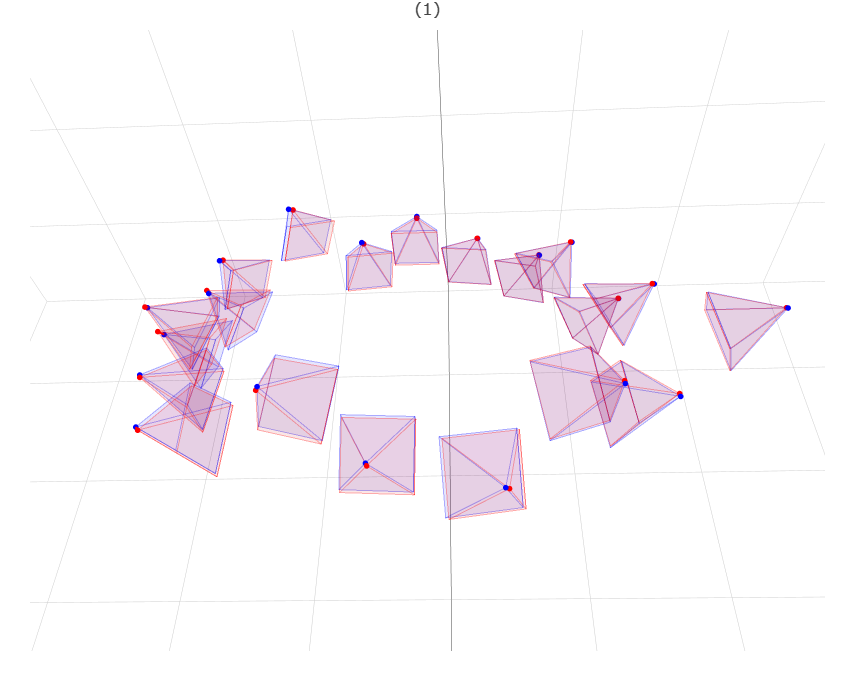}
         &\includegraphics[width = \imagesize\linewidth,  trim = 0pt 0pt 0pt 20pt, clip]{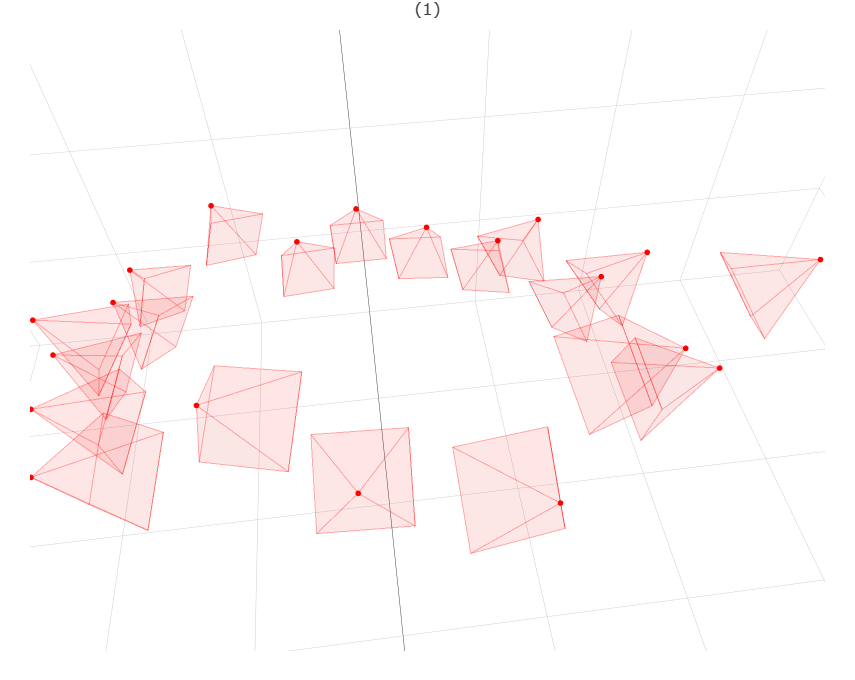}
         \\
               Init. Pose  &  \dust  & Ours & Calib. Pose
    \end{tabular}
	\caption{\textbf{(Top row)} Given multi-view surface normals of a reflective and textureless surface, our method jointly recovers a high-fidelity surface \textbf{(middle row)} and accurate camera poses \textbf{(bottom row)}. The reconstructed shape is comparable to the results of~\cite{supernormal}, which uses calibrated poses.}
	\label{fig:teaser}
    \vspace{-1em}
\end{figure}

Detailed 3D reconstruction from multi-view image observations is a fundamental task in computer vision, empowering various applications like virtual reality and e-Heritage. 
A typical pipeline first calibrates the camera poses for each image and then uses the posed images to recover the shape.
Many methods have achieved promising results in scenes with diffuse and specular surfaces~\cite{neus,nero,refneus,nersp,rnbneus,neisf,underwater}. However, surface reconstruction without pose calibration, which is desired for practical causal capture scenarios, still remains challenging for reflective and textureless surfaces, as shown in \fref{fig:teaser}.

\begin{figure}
    \centering
    \includegraphics[width=\linewidth]{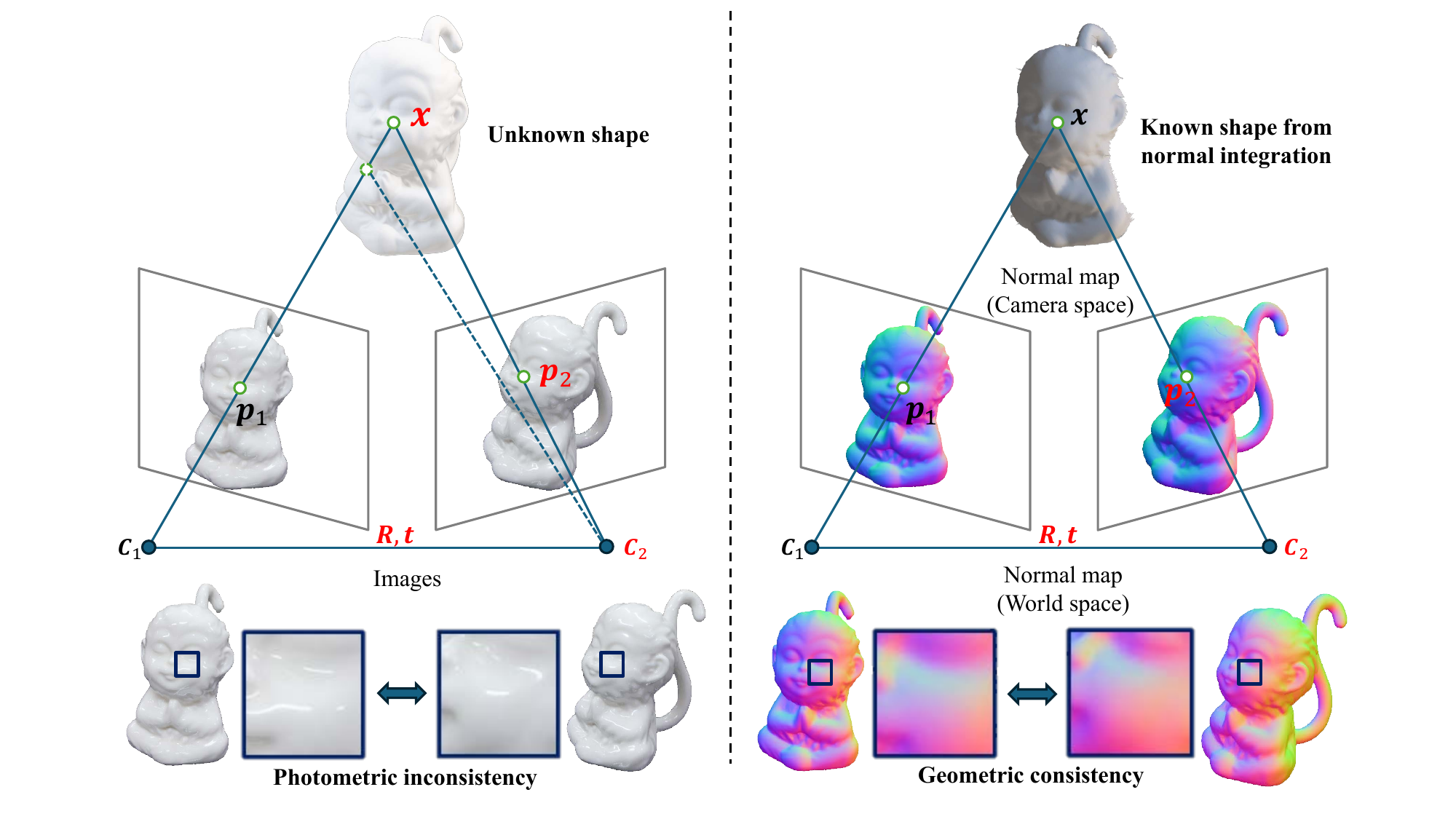}
    \caption{Illustration of shape and pose estimation for reflective and textureless objects based on RGB and surface normals.}
    \label{fig:analysis}
\end{figure}

Existing methods for reconstructing reflective and textureless surfaces, such as NeRO~\cite{nero}, require precise camera pose calibration. 
To achieve this, a calibration board is often placed under the object, limiting the method’s applicability in more casual setups. However, jointly recovering camera poses and surface from images presents a chicken-and-egg problem.
As shown in \fref{fig:analysis}, without knowing the relative pose between cameras $\textbf{c}_1$ and $\textbf{c}_2$, the epipolar plane $\textbf{c}_1-\textbf{c}_2-\V{x}$ remains ambiguous. To mitigate this ambiguity, existing methods either attempt to establish feature correspondences $[\V{p}_1, \V{p}_2]$~\cite{barf, nopenerf} between views or to constrain the shape at specific locations using monocular depth estimation~\cite{cfgs, dust3r}. However, establishing reliable feature correspondences for reflective surfaces is particularly challenging due to view-dependent reflectance. Moreover, the lack of texture further complicates shape estimation using learning-based monocular depth estimators~\cite{depthanything, dpt}. Consequently, there remains a need for a robust 3D reconstruction method that can accurately handle reflective and textureless surfaces while being tolerant to noisy camera poses.

In this paper, we propose Pose-free Multi-view Normal Integration (PMNI), a method that leverages multi-view surface normal maps as input to jointly optimize both surface shape and camera poses. Our key insight is that monocular normal estimation is independent of camera poses, and normal maps encode rich shape information that aids in camera pose estimation. As shown in \fref{fig:analysis}, a normal map can be estimated by photometric stereo from single-view image observations~\cite{sdm}, and is robust to textureless and reflective surfaces. By applying the normal integration method~\cite{bini}, relative depth maps with fine-grained details can be recovered from single-view surface normal maps, providing reliable geometric cues that facilitate camera pose estimation. Moreover, unlike RGB images, where photometric consistency is often disrupted for reflective surfaces, surface normals remain geometrically consistent at corresponding points, making them invariant to changes in surface reflectance.

Building on these insights, we propose a pose-free reflective surface reconstruction method based on multi-view surface normal maps. Specifically, we utilize a signed distance function (SDF) represented by a coordinate-based MLP network, which can simultaneously model both surface shape and surface normals through its analytical gradient. We use per-view depth map, integrated from the surface normal map~\cite{bini}, as an anchor to regularize the SDF. 
At each iteration, with the estimated shape and camera poses, we find correspondences by projecting sampled rays onto the image plane. This allows us to further constrain both the shape and poses by enforcing the geometric consistency of the input surface normals at their projected positions.

As shown in \fref{fig:teaser}, \ours enables the joint recovery of high-fidelity 3D shapes and camera poses, yielding results comparable to methods with calibrated camera poses~\cite{supernormal}, and outperforming existing pose-free 3D reconstruction approaches~\cite{dust3r}.
In this way, \ours makes it possible for reflective and textureless surface reconstruction in a causal capture setting.

\paragraph{Contributions.} This paper proposes \ours, the first method to achieve high-quality reflective surface reconstruction without camera pose calibration. Unlike RGB images, surface normal maps derived from photometric stereo are invariant to reflective and textureless surfaces. We demonstrate normal maps provide an effective regularization for surface reconstruction through integrated depth, effectively reducing ambiguities in both shape and pose recovery. Experiments on both public and our own captured real-world datasets validate the effectiveness of our method.

\section{Related works}
\label{sec:related_works}

This paper focuses on pose-free reflective surface reconstruction from multi-view surface normal maps. 
In the following, \cref{sec.related_reflective} summarizes related works on reflective surface reconstruction that take image observations or surface normal maps as input. 
\Cref{sec.related_pose_free} then surveys pose-free neural radiance field~(NeRF) methods.

\begin{figure}
	\includegraphics[width=\linewidth]{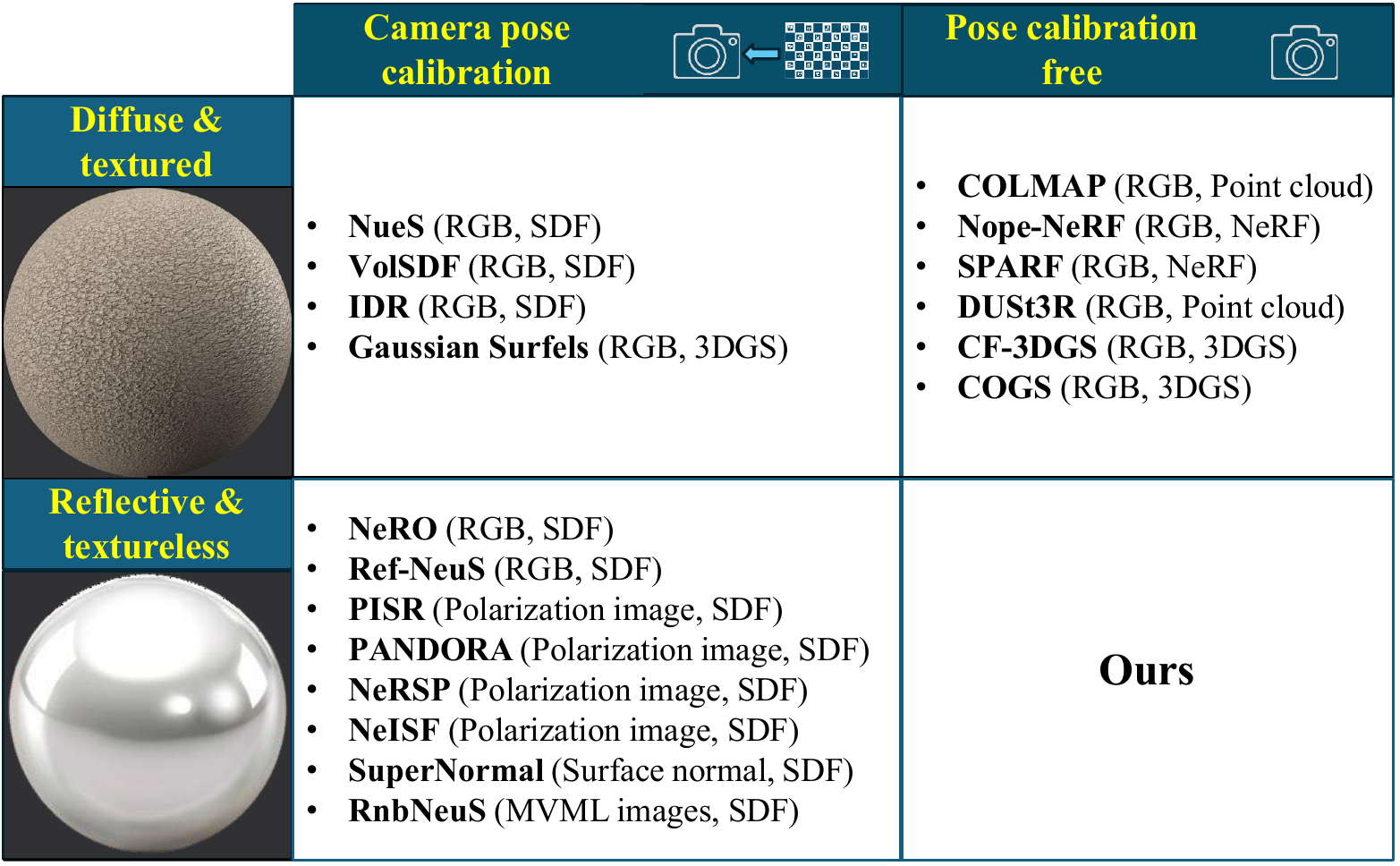}
	\caption{Summary of existing neural surface reconstruction methods categorized by their surface reflectance types and camera calibration settings. The input and surface representation for each method are labeled in brackets.}
	\label{fig:summary_relatedworks}
\end{figure}

\subsection{Neural reflective surface reconstruction}
\label{sec.related_reflective}

Neural 3D reconstruction has advanced significantly since NeRF~\cite{nerf}. Given multi-view images, camera poses are estimated via structure-from-motion through feature matching. Shape represented by SDF~\cite{neus} or Gaussian surfels~\cite{Dai2024GaussianSurfels} is then optimized with differentiable volume rendering.

Reflective surfaces pose additional challenges due to view-dependent reflections, as shown in \fref{fig:summary_relatedworks}. Methods like \nero and \refneus effectively address these issues by using RGB inputs and incorporating Integrated Positional Encoding (IDE) and split-sum approximations to model reflective appearance under environmental lighting.

Polarization-based neural reconstruction, such as \pandora, \nersp, and \pisr, uses the polarization characteristics of diffuse and specular reflectance to address shape-reflectance ambiguity. By decomposing radiance into diffuse and specular components via Stokes vector, these methods improve reflective surface reconstruction. \nersp further integrates geometric cues from the angle of polarization with photometric cues from Stokes vectors, enabling shape estimation even with sparse views.

Photometric stereo excels in reconstructing single-view shapes with complex reflectance by taking images under varying lighting as input. Multi-view photometric stereo (MVPS) extends this approach by combining multi-view, multi-light observations. Methods like \supernormal and \rnbneus first extract per-view normal maps using techniques like \sdm, then refine SDF to align with these normals. Compared to RGB- or polarization-based methods, MVPS is particularly effective for reflective surfaces due to the detailed geometric information encoded in surface normal maps~\cite{supernormal}.

As shown in \Tref{fig:summary_relatedworks}, feature extraction for reflective and textureless surfaces is highly challenging, making camera pose calibration with \colmap unreliable. Consequently, existing reflective surface reconstruction methods rely on a checkerboard during capture, which limits their applicability in casual capture settings. In contrast, PMNI leverages multi-view surface normal maps as input to achieve detailed reflective surface reconstruction without requiring pose calibration.

\subsection{Pose-free surface reconstruction}
\label{sec.related_pose_free}

Given multi-view images, \colmap uses Structure from Motion (SfM) to reconstruct camera poses and sparse 3D points from feature correspondences. For reflective surfaces, where feature matching is challenging, adding a checkerboard can improve reliability.

To address pose errors in \colmap, pose-free methods have been developed to jointly recover surface shapes and camera poses. As summarized in \Tref{fig:summary_relatedworks}, \barf optimizes poses and NeRF using coarse-to-fine positional encoding but requires pose initializations close to the ground truth. \cfgs mitigates this by enforcing temporal continuity and using explicit point cloud representations, though it is limited to dense video sequences.

Pose-free shape reconstruction from sparse views often incorporates learning-based priors. \sparf and \dust rely on pre-trained networks to establish dense 2D correspondences or 2D-to-3D mappings. \cogs and \nopenerf
use monocular depth estimators (\eg, Marigold~\cite{marigold} and DPT~\cite{dpt}) to assist in shape estimation without camera poses. However, these approaches struggle with reflective surfaces.

In summary, pose-free methods often depend on specific initializations, pose continuity, or learning-based priors, which are less effective for reflective surfaces. Additionally, NeRF-based or 3D Gaussian Splatting-based methods often yield noisy shapes. In contrast, PMNI uses SDF representation and surface normal input to achieve detailed shape reconstruction without precise pose initialization.

\section{Proposed method}
\label{sec:method}

We aim to jointly recover fine-grained shapes and camera extrinsic parameters from (1) multi-view camera-space normal images, (2) the corresponding foreground masks of the target object, and (3) camera intrinsic parameters.
To this end, we first perform monocular normal integration on per-view normal maps to obtain per-view relative depth maps; then the normal and depth maps are used together to guide camera pose and shape optimization.

\subsection{Preliminaries}

\paragraph{SDF-based neural surface reconstruction.} 

Signed Distance Function~(SDF) is a common implicit representation of the 3D shape. The surface of the object $\mathcal{M}$ can be viewed as the zero-level set of the SDF:

\begin{equation}
\label{eq.sdf}
 \surface = \{\point \mid f(\point)=0\}.
\end{equation}

Based on \neus, implicit representation of SDF is connected with volume rendering. Specifically, given $K$ ordered 3D points $\{\point_i\}_{i=0}^{K}$ on a ray and their SDF values $\{f(\point_i)\}_{i=0}^{K}$, the volume opacity of a point in space is calculated as follows:
\begin{equation}
\label{eq.sdf_opacity}
    \opacity_i = \max\left(\frac{\sigmoid(f(\point_i))-\sigmoid(f(\point_{i+1}))}{\sigmoid(f(\point_i))}, 0\right),
\end{equation}
where $\sigmoid(x)=1 / \left( 1+\exp(-\sharpness x) \right) $ is the sigmoid function with a learnable sharpness \sharpness. The accumulated transmittance $\transmittance_i$ at a point along a ray can be expressed as:
\begin{equation}
\quad \transmittance_i = \prod_{j=0}^{i-1} (1-\opacity_j).
\end{equation}
Following volume rendering, we render the depth, surface normal, and opacity of a pixel \pixel by
\begin{equation}
\hat{z}\left(\pixel \right) = \sum_{i=0}^{K} T_i \alpha_i d_i,
\end{equation}
\begin{equation}
       \hat{\normal}^w(\pixel) = \sum_{i=0}^{K} \transmittance_i \opacity_i \nabla f (\point_i),
\end{equation}
\begin{equation}
\begin{aligned}
       \quad \hat{o}(\pixel) = \sum_{i=0}^{K} \transmittance_i \opacity_i, 
\end{aligned}
\label{eq.vol_rendering}
\end{equation}
where $\nabla f(\point)$ denotes the gradient of SDF, $\hat{\normal}^w$ denotes world-space surface normal.
By supervising the above volume-rendering information, the SDF network can be constrained, thus accomplishing 3D reconstruction.

\paragraph{Single-view normal integration} aims at reconstructing the relative height map from a given normal map. We define surface normal at \pixel as $\normal^c = [n_x, n_y, n_z]^\top$ in the camera space, the gradient field $[p, q]^\top$ under orthographic projection can be extracted as
\begin{equation}
    p = -\frac{n_x}{n_z}, \quad q = -\frac{n_y}{n_z}.
\end{equation}
The normal integration problem can be formulated as minimizing the following functional:
\begin{equation}
    \mathcal{J}(z) = \iint_{\Omega_n} \left( (\partial_u z - p)^2 + (\partial_v z - q)^2 \right) du \, dv,
\end{equation}
where $\partial_u$ and $\partial_v$ denote partial derivatives of the depth function $z : \Omega_n \to \mathbb{R}$ along $u$ and $v$ axes on the image plane.
Based on this optimization, single-view depth can be obtained up to scale under perspective projection~\cite{bini}. 

\subsection{Problem definition}
Given camera intrinsic $\V{K}$ and multi-view surface normal maps in the camera space, the problem we aim to solve can be formulated as:
\begin{equation}
\min \|\mathbf{R}_i \nabla f(\mathbf{x}) - \mathbf{n}^c_i\left( \V{K} ( \mathbf{R}_i \mathbf{x} + \mathbf{t}_i ) \right) \|_2^2,
\end{equation}
where $[\V{R}_i, \V{t}_i]$ denote rotation and translation at $i$-th view, $\pixel  = \V{K} ( \mathbf{R}_i \mathbf{x} + \mathbf{t}_i)$ represents the projected pixel position of $\point$ at the view, and $\normal^c(\pixel)$ denotes the observed surface normal in camera space. $\mathbf{R}_i \nabla f(\mathbf{x})$ rotates the world-space normal $\normal^w = \nabla f(\mathbf{x})$ to the camera space. By minimizing the difference in camera-space surface normal, this paper aims to solve the 3D surface shape represented by SDF, and the multi-view camera poses jointly.

\subsection{Joint optimization of pose and surface}
\ours adopts a hash-encoded SDF network and uses volume rendering to get per-view surface normal and depth in the world space. We set the SDF network parameters and camera poses as learnable variables, which are optimized via the following loss function:
\begin{equation}
\loss =\lambda_0 \loss_{normal} + \lambda_1 \loss_{ni} + \lambda_2 \loss_{c} + \lambda_3 \loss_{eikonal} + \lambda_4 \loss_{mask},
\end{equation}
where $\lambda_{i}$ is the coefficient to balance different loss terms. In the following, we introduce details of these loss terms.
\paragraph{World-to-camera surface normal loss $\loss_{normal}$.} Given world-space surface normal $\normal^w(\pixel)$ projected at pixel location \pixel rendered from SDF network, and the camera-space surface normal $\normal^c(\pixel)$ recorded in the input surface normal map, $\loss_{normal}$ is defined as
\begin{equation}
\mathcal{L}_{normal} = \sum_{i=1}^N \sum_{\pixel} \left| \V{R}_i\normal_i^w(\pixel) - \normal_i^c(\pixel) \right|_2^2,
\end{equation}
where $N$ denotes the number of input views.

\paragraph{Normal integration loss $\loss_{ni}$.} Given surface normal maps, we use normal integration method
BiNI~\cite{bini} to get integrated depth map $\V{z}^{ni}$. This depth map has an inherent scale ambiguity to the corresponding GT depth, \ie, $\V{z} = \alpha\V{z}^{ni}$. Given the depth map $\V{z}^{r}$ rendered from SDF, we can calculate this scale $\alpha$ via least squares. Specifically,
\begin{equation}
    \alpha = \frac{\V{z}^{ni} \cdot \V{z}^{r}}{\V{z}^{r} \cdot \V{z}^{r}}.
\end{equation}
We calculate the scale for each view. Using these integrated depth maps, we regularize the SDF network by L1 loss, \ie,
\begin{equation}
\mathcal{L}_{ni} = \sum_{i=1}^N \left| \V{z}^{r}_i - \alpha_i\V{z}^{ni}_i \right|.
\end{equation}

\paragraph{Multi-view normal consistency loss $\loss_c$.}
Motivated by existing pose-free 3D reconstruction methods that apply correspondences between views to regularize the camera poses, we try to find dynamic 2D correspondences by projecting sampled scene points from SDF to the 2D image plane defined by camera poses at each iteration. Given these 2D correspondences, we measure the consistency of the surface normal maps in the camera space. 

Specifically, we first cast a ray passing through \pixel at a reference view and trace until touching the surface points \point .

After that, we project \point to all the other camera views via projection $\mathrm{\Pi} = \{\pi_i = [R_i,t_i] \mid i = 0, \dots, N-1 \}$, and get the corresponding surface normals $\normal^{c}(\pi_i(\point))$ in the camera space. Theoretically, these surface normals can be rotated to the same world surface normal via the corresponding camera poses. Based on this constraint, we define the loss at \pixel as
\begin{equation}
\mathcal{L}_{c} = \sum_{i}^{N-1} \left\| \bar{\V{R}} \bar{\normal}^{c}(\pixel) - \V{R}_i \normal^{c}_i(\pi_i(\point))\right\|_2^2,
\end{equation}
where $\bar{\V{R}}$ and $\bar{\normal}$ denote the rotation and camera view normal in the reference frame. 
As point \point is not visible to all views, we introduced a visible mask function $\gamma_i(x) $ based on ray tracing, \ie, 
\begin{equation}
\gamma_i(\mathbf{x}) = 
\begin{cases} 
1 & \text{if } \mathbf{x} \text{ is visible to } i\text{-th camera}, \\
0 & \text{otherwise}.
\end{cases}
\end{equation}
Based on this visibility check, we rewrite the multi-view normal consistency loss as
\begin{equation}
\mathcal{L}_{c} = \sum_{i}^{N-1}\gamma_i(\mathbf{x}) \left\| \bar{\V{R}} \bar{\normal}^{c}(\pixel) - \V{R}_i \normal^{c}_i(\pi_i(\point))\right\|_2^2.
\end{equation}
We compute and accumulate this loss under different pixels and reference views.

\paragraph{Mask loss $\loss_{mask}$} is built upon labeled silhouette of target shape, \ie,  
\begin{equation}
\mathcal{L}_{mask} = \sum_i^N\sum_{\pixel} \bce\left(\hat{o}_i(\pixel), o_i(\pixel)\right),
\end{equation}
where $o_i(\pixel)$ and $\hat{o}_i(\pixel)$ correspond to the input and rendered mask value at \pixel, and $\bce(\cdot)$ denotes binary cross entropy function.

\paragraph{Eikonal loss $\loss_{eikonal}$.}
To enforce the SDF gradient norm to be close to $1$ almost everywhere so that the neural SDF is approximately valid, we introduce Eikonal loss as follows,
\begin{equation}
\mathcal{L}_{eikonal} = \sum_{\point} \left(\norm{\nabla f(\point)}_2 - 1\right)^2.
\end{equation}

\paragraph{Initialization of camera poses.}
We initialize the camera poses as a circular distribution with a radius $r$. To determine $r$, we assume the target object is within a bounding box $[-1, 1]^3$, and the object is always fully covered by each view with the resolution of $H \times W$. Given focal length in pixel units, radius $r$ can be determined by $2f / H$. More details can be found in our supplementary material.

\section{Experiment} \label{sec:exp}

In this section, we first evaluate the shape estimation of our method using multi-view normal integration with calibrated camera poses as reference~(\cref{sec.benchmark_eval}). Then, we compare pose-free 3D reconstruction methods with ours on both pose and shape estimation~(\cref{sec.pose_free_eval}). More experiments, such as shape reconstruction from sparse and uncalibrated camera poses, are in the supplementary material.

\begin{table}
\caption{Quantitative evaluation of shape and camera pose recovery on \dmv. \supernormal with noisy camera poses is indicated with * marker. The best and second-best results are labeled in \textbf{bold} and \underline{underlined}.}
	\resizebox{\linewidth}{!}{
	\begin{tabular}{@{}l|c|ccccc|c@{}}
		\toprule
		 Method & Metric & {\sc\small Bear} & {\sc\small Buddha} & {\sc\small Cow} & {\sc\small Pot2} & {\sc\small Reading}  & Average \\
		\midrule
        
		\supernormal & \multirow{3}{*}{CD~$\downarrow$} & \textbf{0.158} & \textbf{0.111} & \textbf{0.099} & \underline{0.154} & \underline{0.187} & \textbf{0.142} \\
        
		\supernormal* && 0.614 & 0.862 & 0.985 & 0.771 & 0.645 & 0.775 \\
        
		\ours && \underline{0.189} & \underline{0.122} & \underline{0.191} & \textbf{0.115} & \textbf{0.148}  & \underline{0.153} \\

		\midrule
        
		\supernormal & \multirow{3}{*}{F1-score~$\uparrow$} & \textbf{0.982} & \textbf{0.998} & \textbf{0.999} & \underline{0.998} & \underline{0.988}  & \textbf{0.993} \\
        
        \supernormal* && 0.500 & 0.356 & 0.310 & 0.421 & 0.465  & 0.410 \\
        
		\ours && \underline{0.970} & \underline{0.996} & \underline{0.989} & \textbf{0.999} & \textbf{0.995}  & \underline{0.990} \\

        \midrule
        \multirow{2}{*}{PMNI} & RPEr($^\circ$)~$\downarrow$ & 0.115 & 0.231  & 0.184 & 0.141 & 0.209 & 0.176\\
			 & RPEt~$\downarrow$ & 0.030 & 0.059  & 0.044 & 0.037 & 0.087 & 0.051 \\

		\bottomrule
	\end{tabular}
	}
	\label{tab:dmv_quan_eval}
\end{table}

\renewcommand{\imagesize}{0.25}
\begin{figure}

    \includegraphics[width=\linewidth]{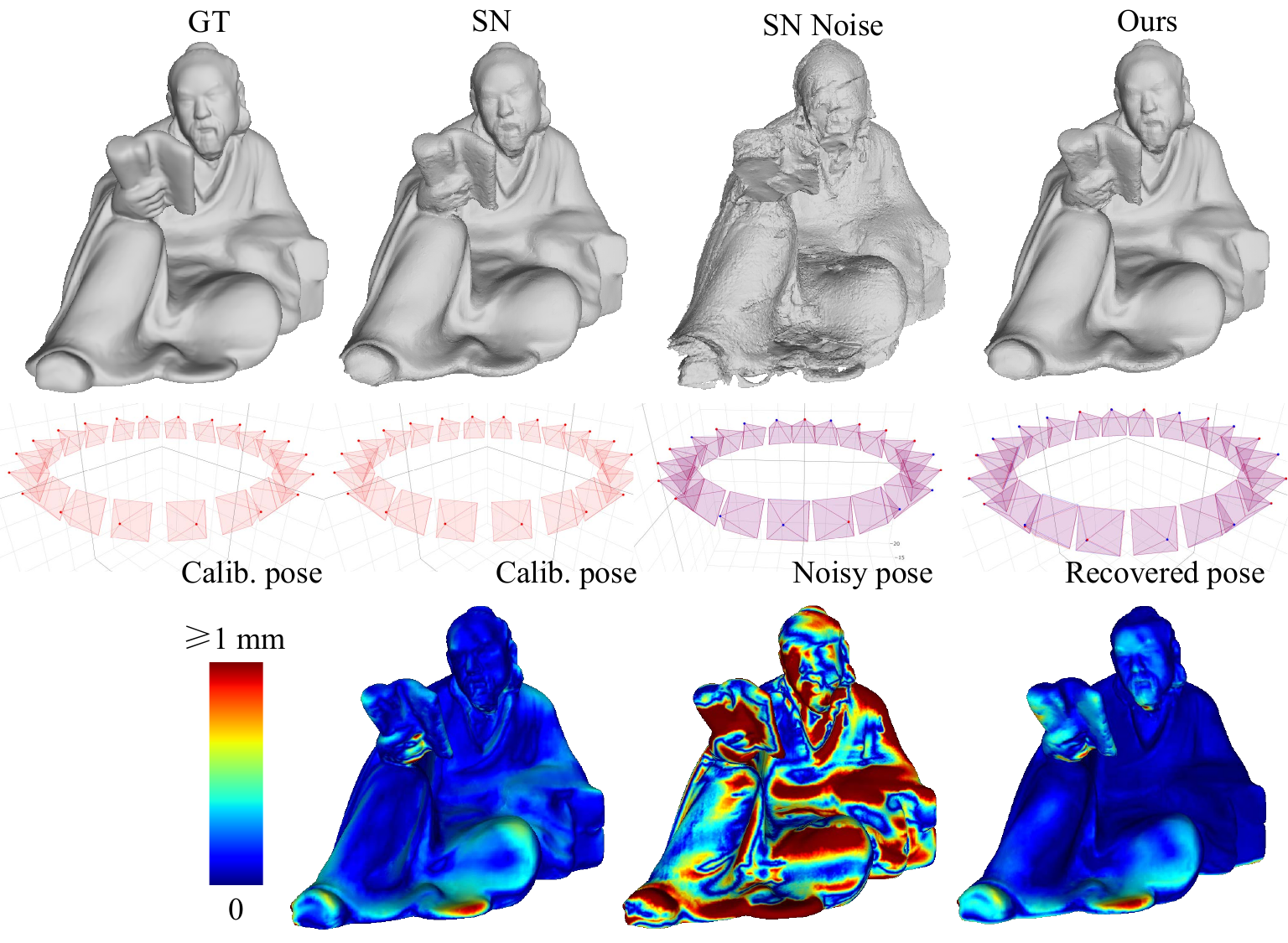}
    \includegraphics[width=\linewidth]{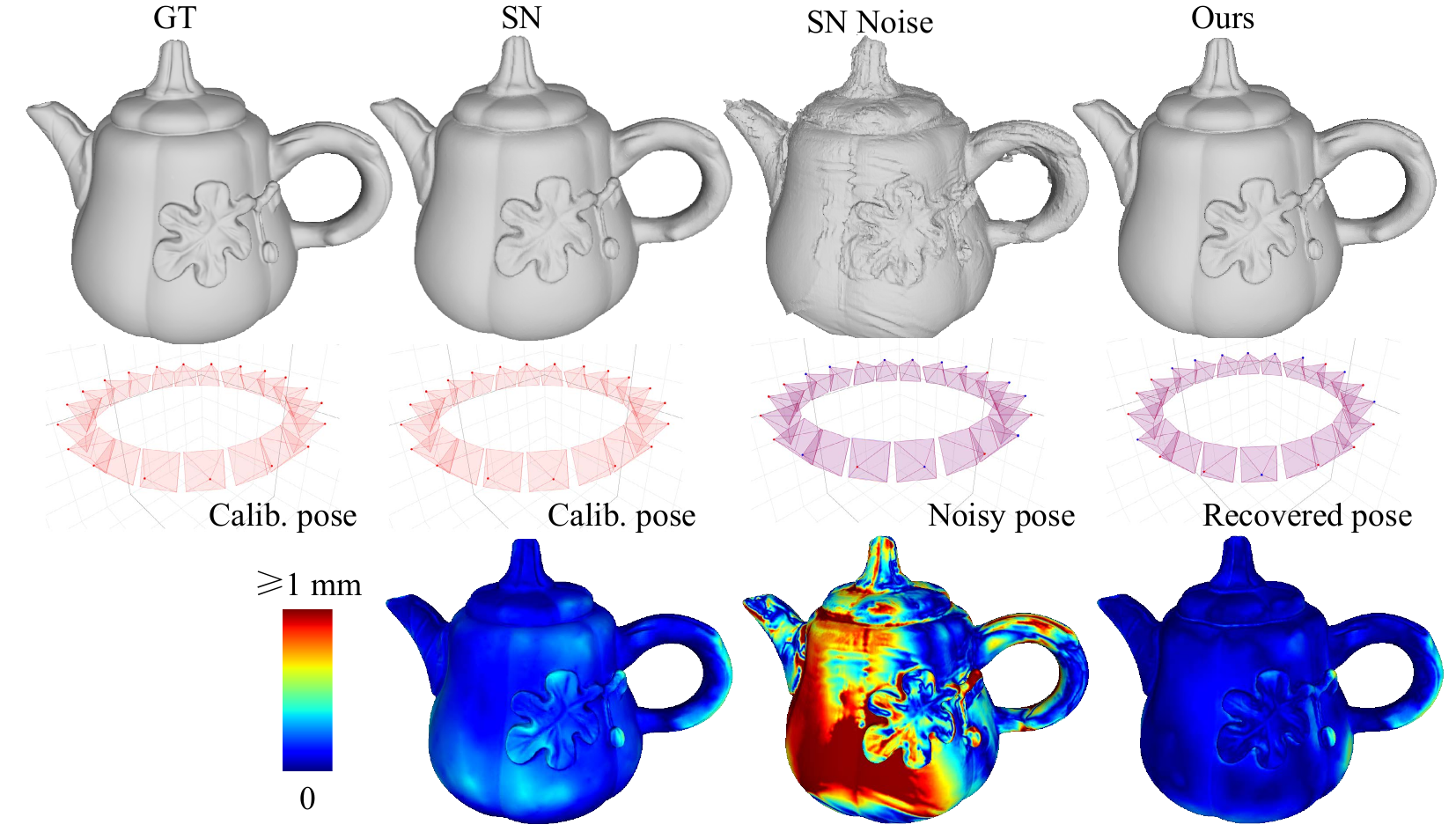}
    
	\caption{Qualitative comparison between \supernormal~(abbreviated by SN) and ours on \dmv. The camera poses for ``SN noise'' are slightly perturbed to simulate calibration noise. Our method accurately recovers camera poses, and the reconstruction is robust to pose calibration noise.}
	\label{fig:dmv_qual_shape}
\end{figure}

\subsection{Comparison on multi-view normal integration}
\label{sec.benchmark_eval}

\paragraph{Dataset and baselines.}
\dmv includes $5$ objects captured from $20$ views, providing ground-truth 3D meshes, per-view surface normals, and calibrated camera poses. Using \dmv, we compare our method with the state-of-the-art multi-view normal integration method \supernormal, which uses calibrated camera poses as input.

\paragraph{Evaluation metric.}
We evaluate shape accuracy using the L2 Chamfer distance~(CD) and F-score with a threshold of $\tau_F = 0.5 mm$, following \supernormal. For pose estimation, we align the estimated poses to the ground truth using~\cite{umeyama1991least}. Following \nopenerf, Relative Pose Error (RPE) is adopted, consisting of relative rotation error (RPEr) and relative translation error (RPEt) to assess errors between image pairs.

\paragraph{Shape and pose recovery results.}

As shown in \Tref{tab:dmv_quan_eval}, we perform a quantitative comparison between \supernormal and \pmni on \dmv, using the GT surface normal as input. We observe that \pmni achieves comparable 3D shape reconstruction to \supernormal, with better results on the {\sc Reading} object. As shown in \fref{fig:dmv_qual_shape}, the CD error distributions show that both \supernormal and our method recover shapes close to the GT, highlighting the effectiveness of our pose-free multi-view normal integration approach.

We also evaluate \supernormal's robustness to camera pose calibration. Specifically, Gaussian noise with variations of 0.01 for translation and 0.287$^\circ$ for rotation is added to the camera poses. Despite these small perturbations, the recovered shapes exhibit high-frequency artifacts, as shown in \fref{fig:dmv_qual_shape}. This occurs because noisy camera poses lead to inconsistencies in multi-view surface normal projections. In contrast, by jointly optimizing camera poses and surface shapes, our \pmni method is robust to calibration noise and produces significantly smaller shape estimation errors than \supernormal.

\begin{figure}
    \includegraphics[width=\linewidth]{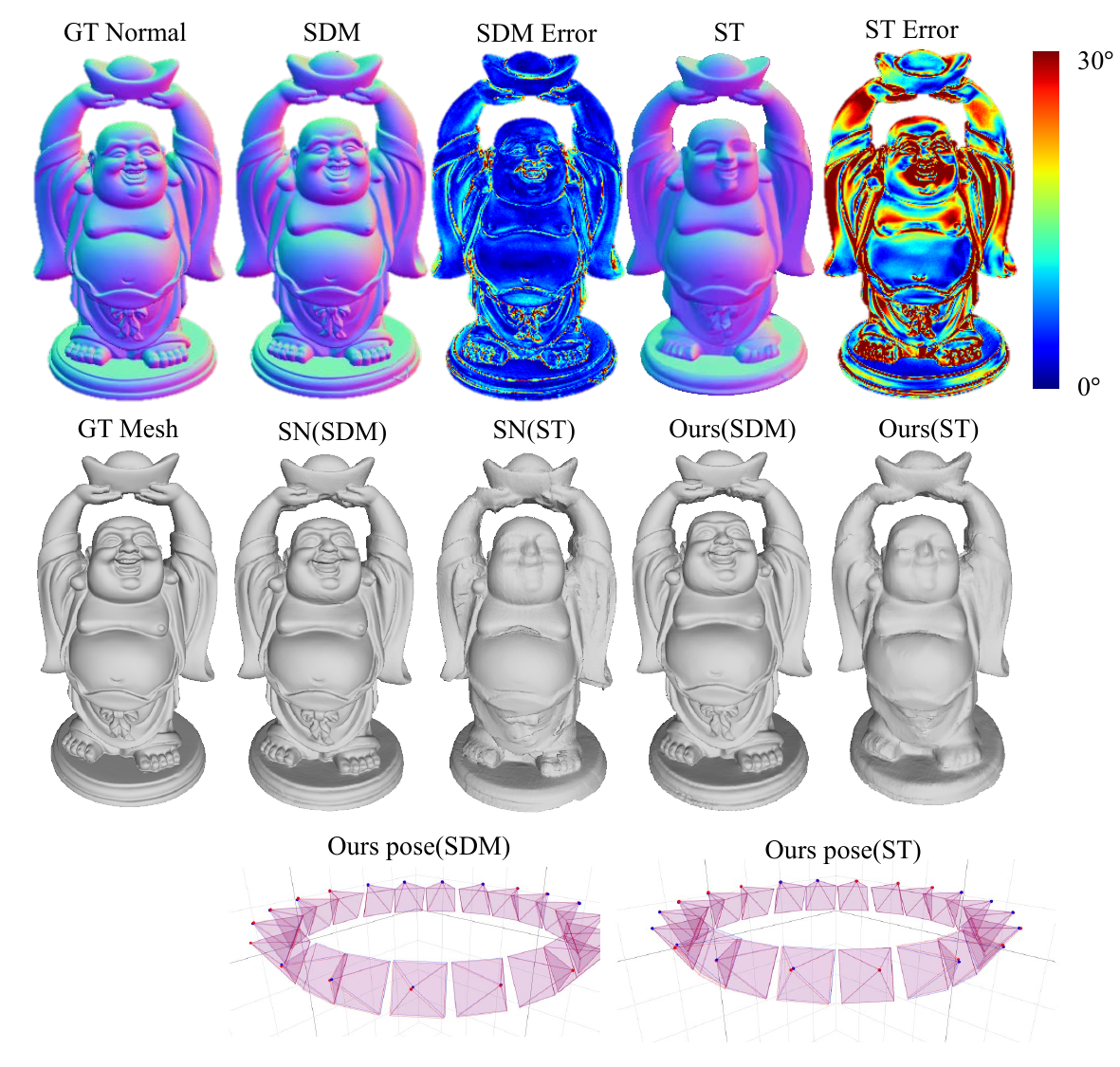}
    \caption{Qualitative comparison with \supernormal ~(abbreviated as ``SN'') using surface normal maps estimated by \sdm ~and \stablenormal (abbreviated as ``SDM'' and ``ST''), respectively. The top row visualizes the input surface normals and their angular error distributions.}
    \label{fig:normal_qual_shape}
\end{figure}

\paragraph{Evaluation under different input surface normals.}
Multi-view normal integration is flexible regarding the source of input surface normal maps. From a practical standpoint, it is important to assess the robustness of \pmni to errors introduced by different normal estimators. Specifically, we use the photometric stereo method \sdm, which relies on images under varying lighting, and the single-image normal estimator \stablenormal to generate input surface normals for both our method and \supernormal. As shown in \fref{fig:normal_qual_shape}, surface normals from \stablenormal are less accurate than those from \sdm. However, \pmni still achieves 3D shape estimations comparable to \supernormal. Despite differences in input, the recovered poses based on \stablenormal or \sdm remain close to the GT.
\Tref{tab:PS_eval} further summarizes shape and pose estimation errors on \dmv, showing that \pmni and \supernormal are comparable, which highlights the robustness of our method to varying levels of input surface normal errors.

\begin{figure*}
    \includegraphics[width=\linewidth]{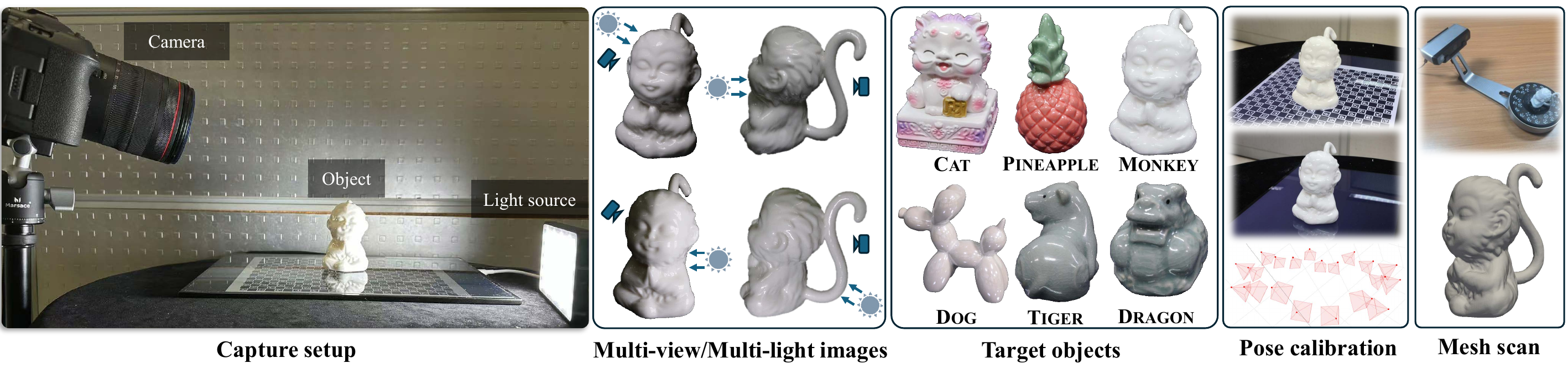}
    \caption{Summary of RT3D dataset for pose-free reflective surface reconstruction. }
    \label{fig:rg10_data}
\end{figure*}

\begin{table}
	\caption{Quantitative evaluation of shape and pose recovery using normal maps from \sdm and \stablenormal. The mean angular errors~(MAE) of the input normal maps are shown in the header.}
	\resizebox{\linewidth}{!}{
		\begin{tabular}{@{}lcccc@{}}
			\toprule
			Method & Metric & \makecell{\stablenormal \\ (MAE: 23.6$^\circ$)} & \makecell{\sdm \\ (MAE: 8.3$^\circ$)} & \makecell{GT Normal\\ (MAE: 0$^\circ$)} \\
			\midrule
			\supernormal & \multirow{2}{*}{CD~$\downarrow$} & 0.543 & 0.194  & 0.142 \\
			\ours & & 0.602 & 0.252  & 0.153 \\
			\midrule
			\supernormal & \multirow{2}{*}{F1-score~$\uparrow$} & 0.644 & 0.962  & 0.993 \\
			\ours & & 0.620 & 0.946  & 0.990 \\
			\midrule
			\multirow{2}{*}{PMNI} & RPEr($^\circ$)~$\downarrow$ & 1.375 & 0.304  & 0.176 \\
			 & RPEt~$\downarrow$ & 0.384 & 0.095  & 0.051 \\
			\bottomrule
		\end{tabular}
	}
	\label{tab:PS_eval}
\end{table}

\subsection{Comparison on pose-free 3D reconstruction}
\label{sec.pose_free_eval}
This section evaluates previous pose-free surface reconstruction methods on reflective and textureless objects. 

 \paragraph{Baselines.} We select \dust, \sparf, \nopenerf, and \cfgs as baselines for pose-free 3D reconstruction. Our experiments show that \sparf and \nopenerf are sensitive to pose initialization. Therefore, we initialize their poses with calibrated values while allowing them to be learned during optimization. In contrast, our method initializes camera poses in a circular distribution, as detailed in the supplementary material. \dust and \cfgs do not require pose initialization.

Since \sparf, \nopenerf, and \cfgs focus on pose-free novel view synthesis and output per-view depth for geometric estimation, we evaluate them quantitatively using depth maps. However, depth maps $\hat{\V{z}}$ from both existing methods and ours have global scale ambiguity compared to the GT. We use the depth map $\V{z}_s$ from \supernormal with calibrated camera poses as the GT reference and compute a global scale $s$ that minimizes the difference between $s \hat{\V{z}}$ and $\V{z}_s$ using least squares. The relative depth error is then defined as the mean absolute difference between the scaled depth $s \hat{\V{z}}$ and $\V{z}_s$ divided by $\V{z}_s$.

\paragraph{\rgreal dataset.} To quantitatively evaluate reconstruction quality on reflective surfaces, we construct a multi-view dataset with ground-truth meshes. \cref{fig:rg10_data} shows our captured $6$ objects with highly reflective surfaces. For each object, we use a Canon EOS R5 camera to capture $20$ views surrounding the object. For each view, we take $11$ images under varying illumination by moving an area light source to different positions. These multi-light images are used for photometric stereo to generate reliable surface normals.

To facilitate camera pose calibration, we place each target object on an OLED screen displaying ArUco markers, as shown in \cref{fig:rg10_data}. The scene is captured twice, once with the display on and once off. The images with ArUco markers are used for evaluation only. Images without ArUco markers serve as input for baseline methods and our approach. Additionally, we scan the shape of the $6$ objects with an EinScan SP scanner\footnote{\url{https://www.einscan.com/einscan-sp}. Retrieved Nov. 14th, 2024.}, which provides a reference for qualitatively assessing the reconstructed shapes.

\begin{table}
	\caption{Quantitative comparison between existing methods and ours on camera pose and surface shape estimation.}
	\resizebox{\linewidth}{!}{
		\begin{tabular}{@{}lcccccccc@{}}
			\toprule
            \multirow{2}{*}{Method} & \multicolumn{7}{c}{RPEr($^\circ$)~$\downarrow$}  \\
			& {\sc\small Monkey} & {\sc\small Cat} & {\sc\small Pineapple} &  {\sc\small Dog} & {\sc\small Dragon} & {\sc\small Tiger} & Avg \\
			\midrule
			{\dust} & \underline{3.175} & \underline{2.049} & \underline{2.640} & \underline{2.216} & 2.602 & 4.839 & \underline{2.920} \\

			{\nopenerf} & 9.371 & 8.472 & 7.513 & 8.674 & 8.467 & 8.282 & 8.463  \\
			{\sparf} & 7.233 & 6.395 & 3.485 & 3.620  & \underline{0.731} & \underline{0.695} & 3.693 \\
			{\cfgs} & 16.867 & 16.664 & 17.276 & 14.789  & 15.625 & 16.659 & 16.313 \\
			{\ours} & \textbf{0.230} & \textbf{0.356} & \textbf{0.258} & \textbf{0.258} & \textbf{0.439} & \textbf{0.582} & \textbf{0.354} \\

			\bottomrule
            
			\multirow{2}{*}{Method} & \multicolumn{7}{c}{RPEt~$\downarrow$}  \\
			& {\sc\small Monkey} & {\sc\small Cat} & {\sc\small Pineapple} &  {\sc\small Dog} & {\sc\small Dragon} & {\sc\small Tiger} & Avg \\
			\midrule
			{\dust} & \underline{0.329} & \underline{0.199} & 0.247 & 0.490 & 0.224 & 0.335 & 0.304  \\
			{\nopenerf} & 0.695 & 0.596 & 0.610 & 0.774 & 0.654 & 0.637 & 0.661  \\
			{\sparf} & 0.375 & 0.203 & \underline{0.146} & \underline{0.261} & \underline{0.041} & \underline{0.058} & \underline{0.181} \\

			{\cfgs} & 0.947 & 0.796 & 1.092 & 0.878 & 0.998 & 1.124 & 0.972 \\
			{\ours} & \textbf{0.015} & \textbf{0.020} & \textbf{0.016} & \textbf{0.019} & \textbf{0.027} & \textbf{0.035} & \textbf{0.022} \\

			\bottomrule

            \multirow{2}{*}{Method} & \multicolumn{7}{c}{Relative Depth Error~$\downarrow$}  \\
			& {\sc\small Monkey} & {\sc\small Cat} & {\sc\small Pineapple} &  {\sc\small Dog} & {\sc\small Dragon} & {\sc\small Tiger} & Avg \\
			\midrule
			{\dust} & \underline{0.062} & 0.056 & 0.046 & 0.147 & 0.046 & 0.075 & 0.072 \\
                {\nopenerf} & 0.276 & 0.191 & 0.305 & 0.489 & 0.231 & 0.176 & 0.278 \\
               {\sparf} & 0.099 & \underline{0.055} & \underline{0.038} & \underline{0.131} & \underline{0.029} & \underline{0.050} & \underline{0.067} \\

                {\cfgs} & 0.363 & 0.360 & 0.475 & 0.488 & 0.477 & 0.502 & 0.444 \\
                {\ours} & \textbf{0.011} & \textbf{0.017} & \textbf{0.008} & \textbf{0.010} & \textbf{0.011} & \textbf{0.026} & \textbf{0.014} \\

			\bottomrule
		\end{tabular}
 	}
	\label{tab:synthetic_qual_pose}
\end{table}
\paragraph{Pose evaluation.}
As shown in \fref{fig:real_world_eval}, we visualize the GT~(shown in red) and estimated poses~(shown in blue) from existing methods and ours on {\sc Monkey} and {\sc Dog} object. The red line connecting the GT and estimated camera positions illustrates the performance of pose recovery. \cfgs and \nopenerf cannot produce reasonable pose estimation, possibly due to the temporal continuity assumption, which is not satisfied in the pose distribution of \rgreal. \sparf applies a pre-trained dense correspondence network, which may not generalize well on reflective and textureless surfaces such as {\sc Monkey} and {\sc Dog}, affecting the pose estimation. \dust has relatively better results based on learned point cloud correspondence but there still remains a gap between its poses and GT.
Given a circular pose initialization shown in the second column, the estimated poses from our method are accurately aligned with the corresponding GT, as shown in the last column.

As shown in the top and middle rows of \Tref{tab:synthetic_qual_pose}, our recovered poses, including rotation and translation, achieve state-of-the-art performance over existing methods, demonstrating the strength of using multi-view surface normals for optimizing the camera poses.

\paragraph{Shape evaluation.}
As shown in \fref{fig:shape_est}, we compare estimated shapes from existing methods and ours, where \dust and our method can output multi-view mesh, and the shape visualizations from other methods are based on depth. Consistent with the pose estimation, \dust obtains better results than existing pose-free methods, but is still unsatisfactory compared with scanned meshes. In contrast, our \pmni gets detailed shape recoveries for the two reflective and textureless surfaces, and the results are close to \supernormal and scanned meshes, showing the strength of our method. More results on \rgreal can be found in the supplementary material.

\renewcommand{\imagesize}{0.145}
\begin{figure*}
	\small
    \begin{tabular}{@{}c@{}@{}c@{}@{}c@{}@{}c@{}@{}c@{}@{}c@{}@{}c@{}}
         Calib. pose &Init. pose &\dust & \sparf & \nopenerf & \cfgs & Ours\\
         \includegraphics[width = \imagesize\linewidth,   trim = 0pt 0pt 0pt 80pt, clip]{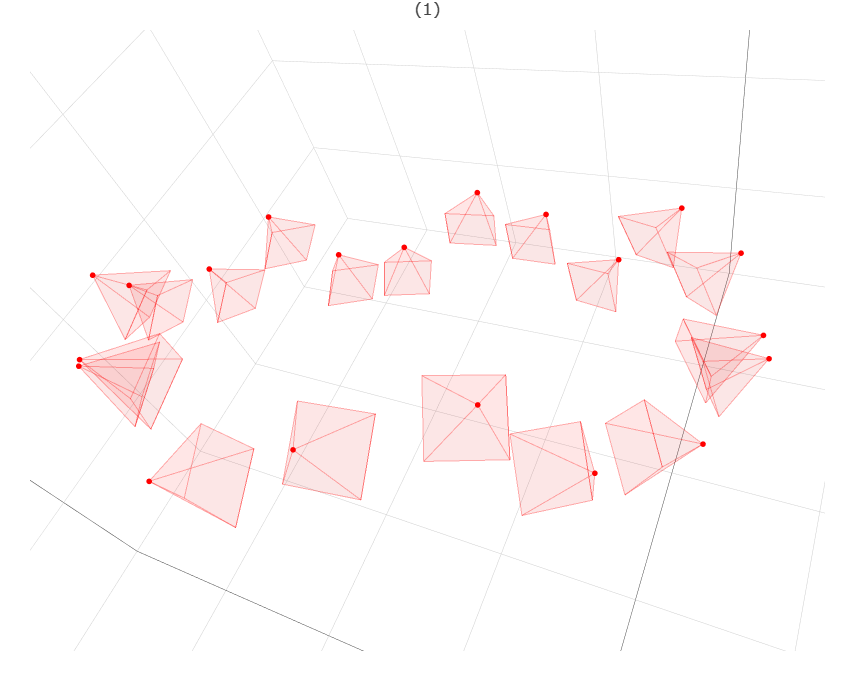}
        &\includegraphics[width = \imagesize\linewidth,  trim = 0pt 0pt 0pt 80pt, clip]{fig/fig1/init.png}
         &\includegraphics[width = \imagesize\linewidth,  trim = 0pt 0pt 0pt 80pt, clip]{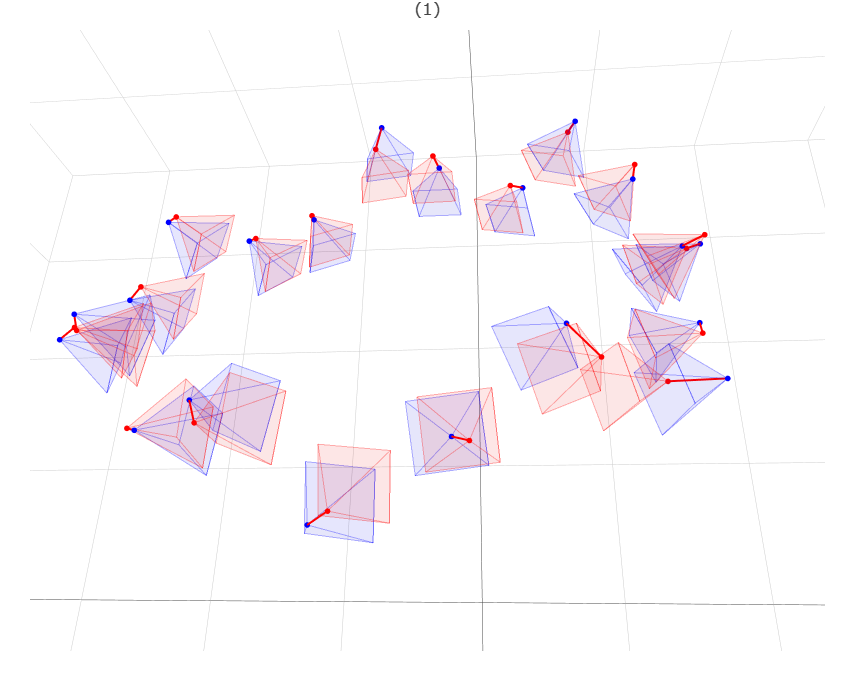}
         &\includegraphics[width = \imagesize\linewidth,  trim = 0pt 0pt 0pt 80pt, clip]{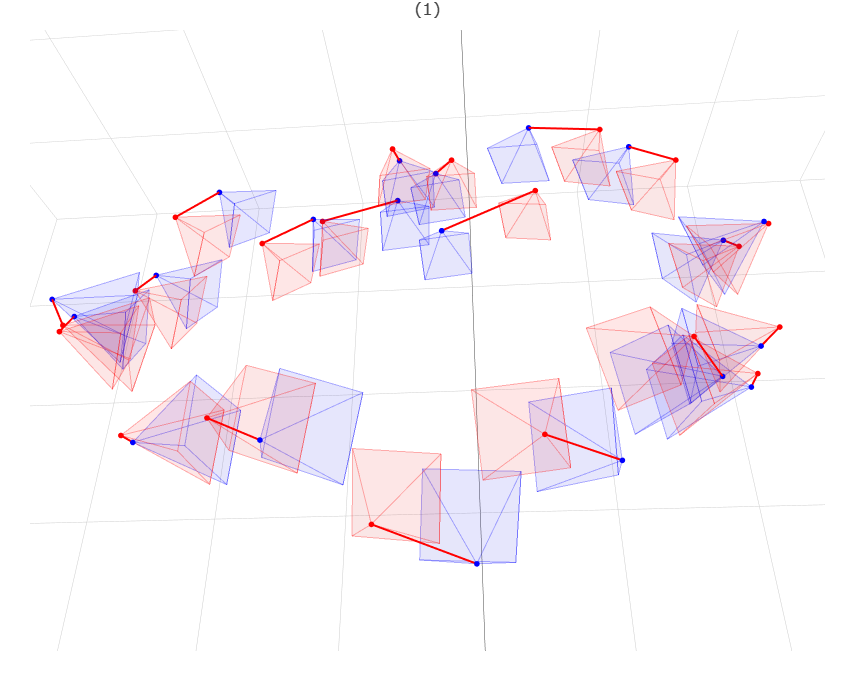}
         &\includegraphics[width = \imagesize\linewidth,  trim = 0pt 0pt 0pt 80pt, clip]{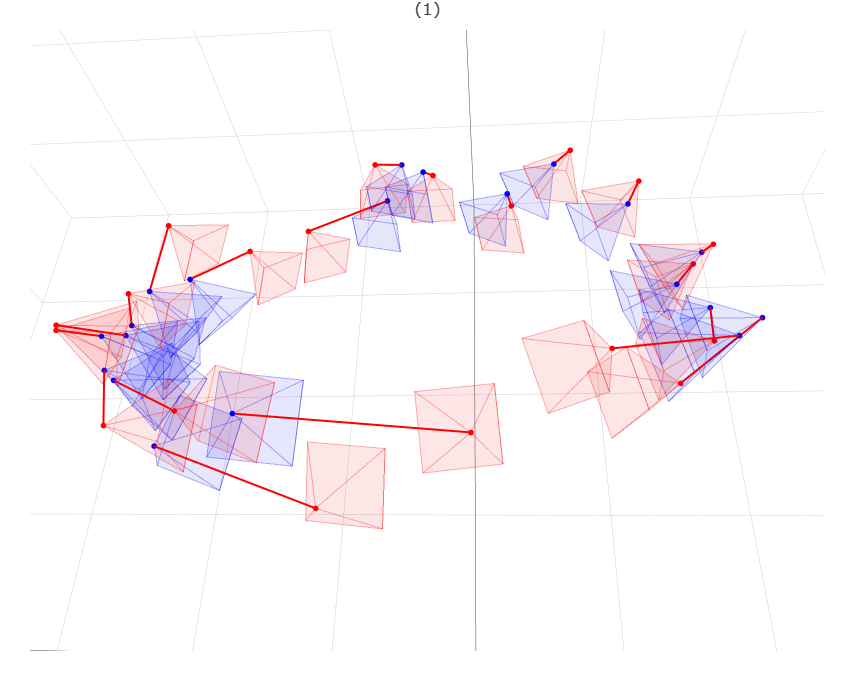}
         &\includegraphics[width = \imagesize\linewidth,  trim = 0pt 0pt 0pt 80pt, clip]{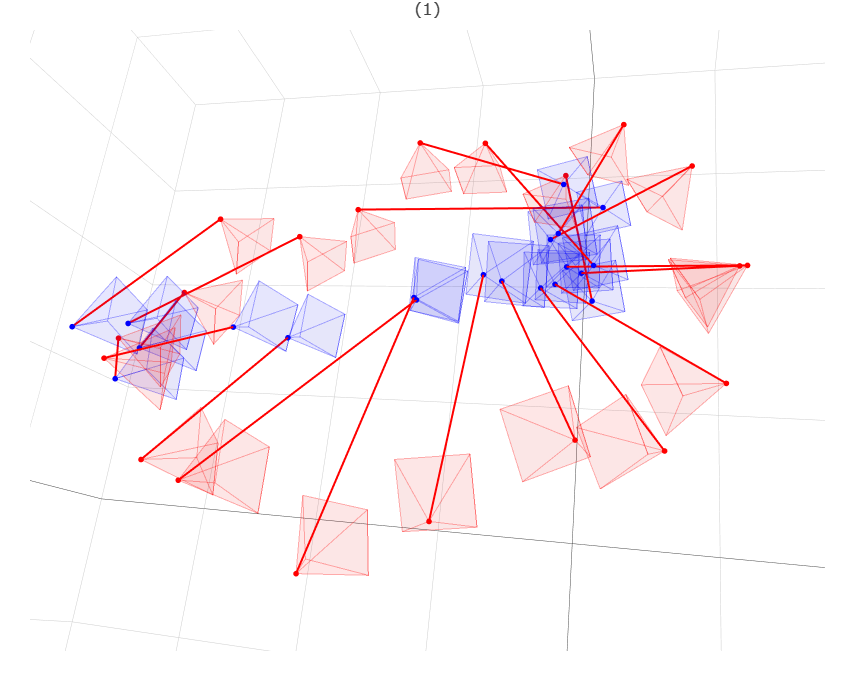}
         &\includegraphics[width = \imagesize\linewidth,  trim = 0pt 0pt 0pt 80pt, clip]{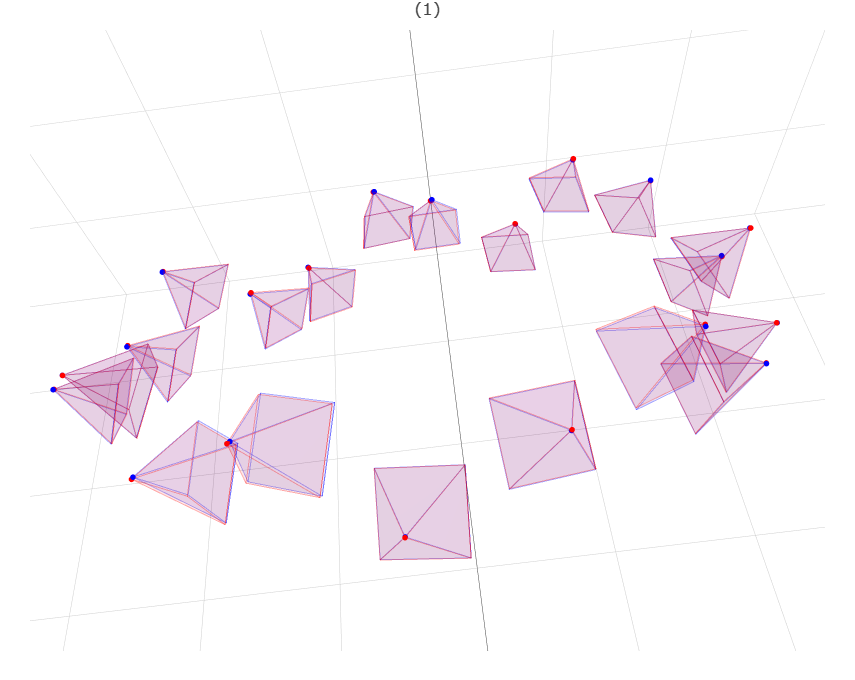}
          \\
        \includegraphics[width = \imagesize\linewidth,  trim = 0pt 60pt 0pt 90pt, clip]{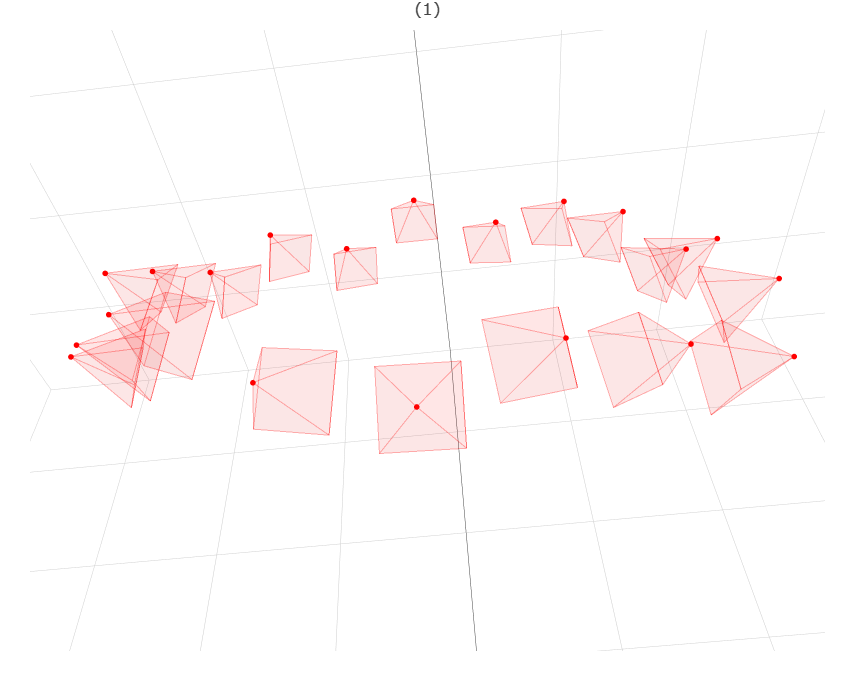}
        &\includegraphics[width = \imagesize\linewidth,  trim = 60pt 60pt 0pt 20pt, clip]{fig/fig1/init.png}
         &\includegraphics[width = \imagesize\linewidth,  trim = 0pt 60pt 0pt 90pt, clip]{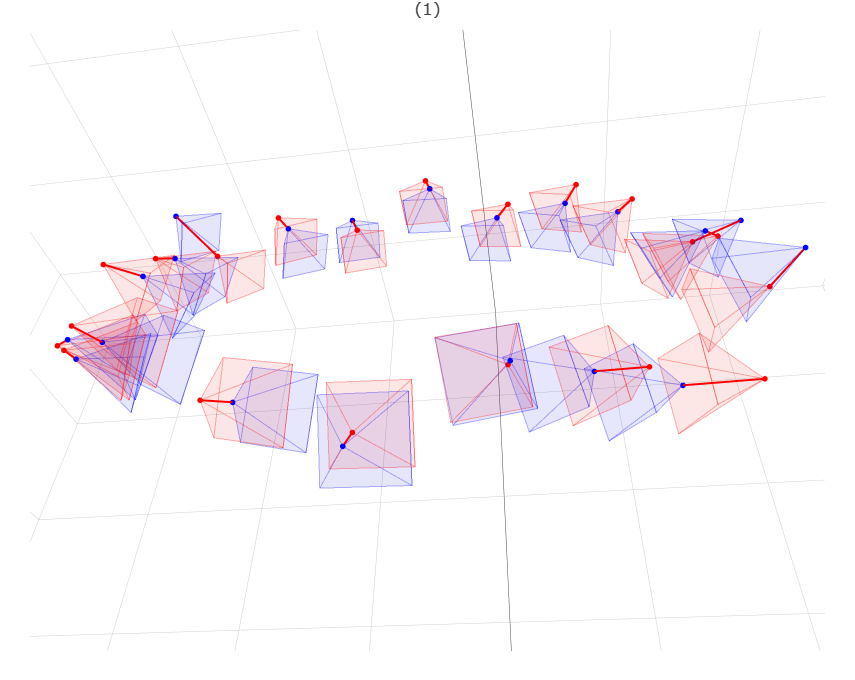}
         &\includegraphics[width = \imagesize\linewidth,  trim = 0pt 60pt 0pt 90pt, clip]{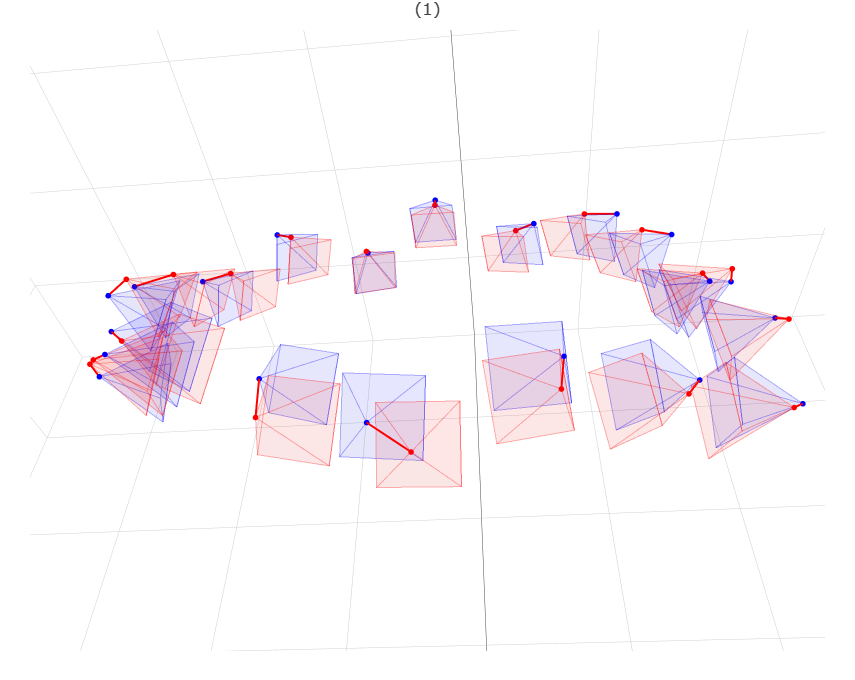}
         &\includegraphics[width = \imagesize\linewidth,  trim = 0pt 60pt 0pt 90pt, clip]{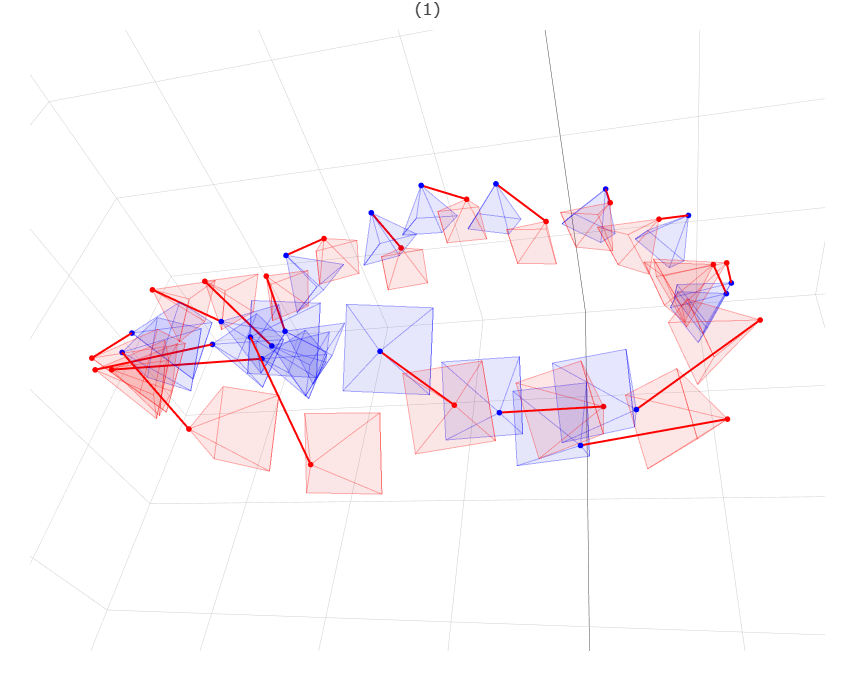}
         &\includegraphics[width = \imagesize\linewidth,  trim = 0pt 40pt 0pt 90pt, clip]{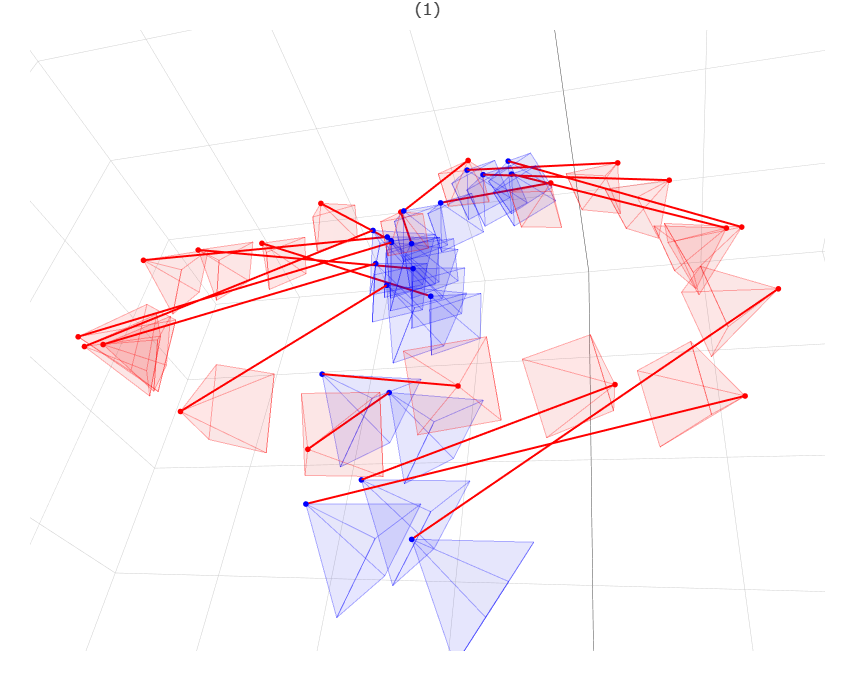}
         &\includegraphics[width = \imagesize\linewidth,  trim = 0pt 60pt 0pt 90pt, clip]{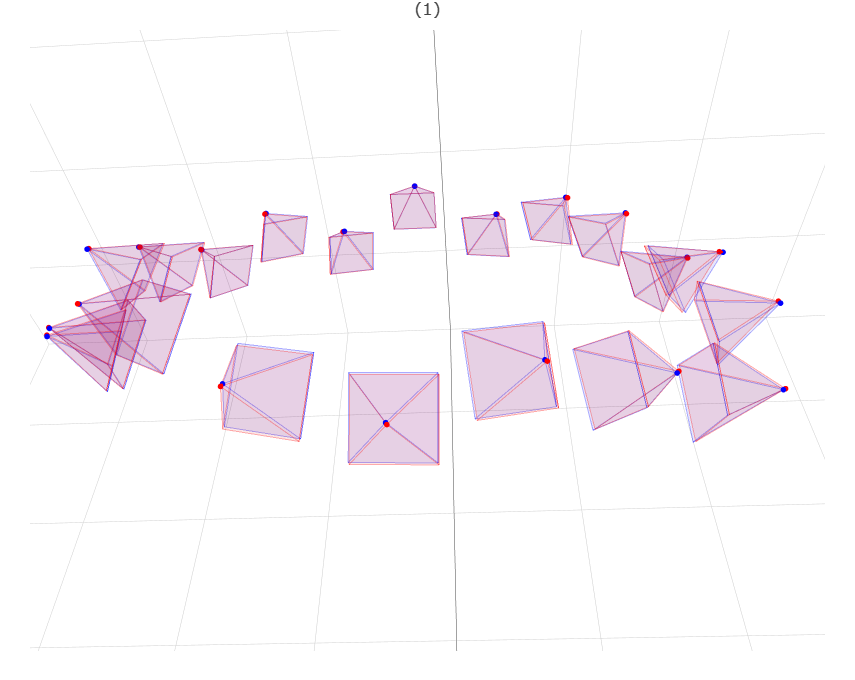}
    \end{tabular}
	\caption{Qualitative comparison of camera pose recovery on {\sc Monkey} and {\sc Dog} object of \rgreal dataset. The red line segment connects the calibrated and estimated camera locations to illustrate the quality of pose recovery. }
	\label{fig:real_world_eval}
\end{figure*}

\renewcommand{\imagesize}{0.1}
\begin{figure*}
\tiny
\resizebox{\linewidth}{!}{
    \begin{tabular}{@{}c@{}@{}c@{}@{}c@{}@{}c@{}@{}c@{}@{}c@{}@{}c@{}@{}c@{}@{}c@{}}
     & GT &\supernormal & \dust & \sparf & \nopenerf & \cfgs & Ours\\
        \includegraphics[width = 0.099\linewidth, trim = 355pt 115pt 340pt 40pt, clip]{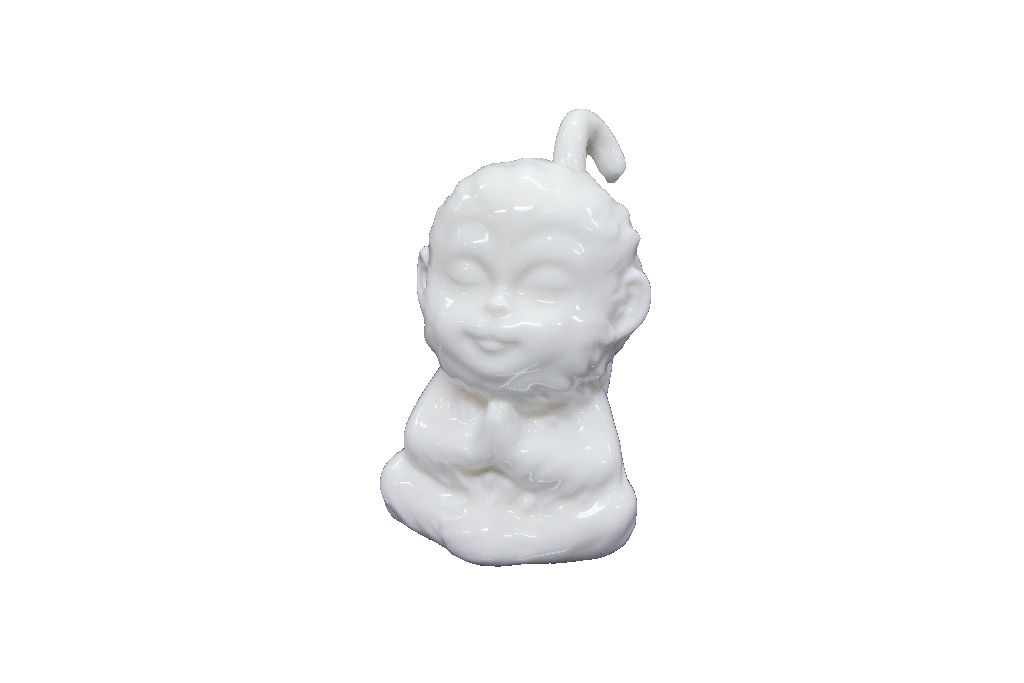}
         &\includegraphics[width = 0.084\linewidth, trim = 0pt 0pt 0pt 0pt, clip]{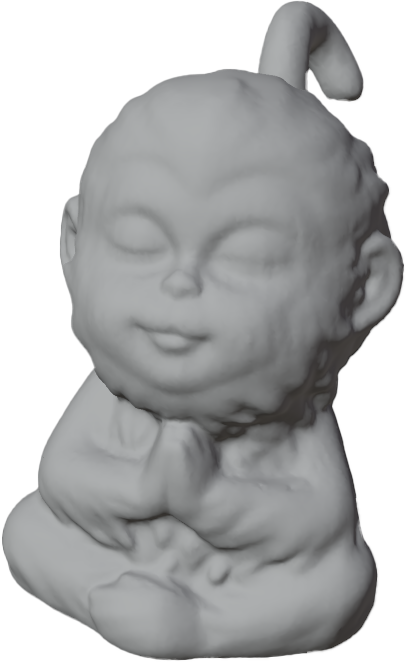}
         &\includegraphics[width = 0.084\linewidth, trim = 0pt 0pt 0pt 0pt, clip]{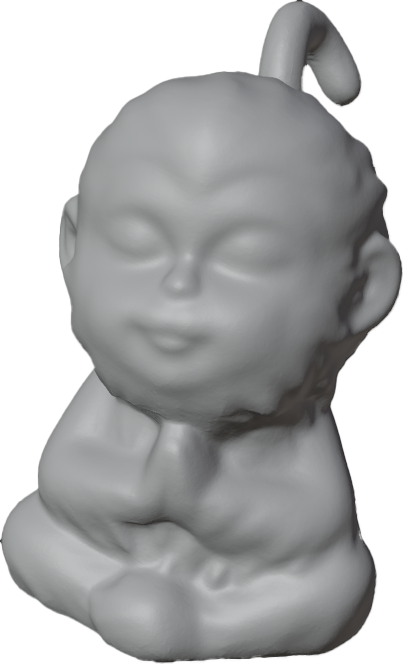}
         &\includegraphics[width = \imagesize\linewidth, trim = 355pt 170pt 340pt 140pt, clip]{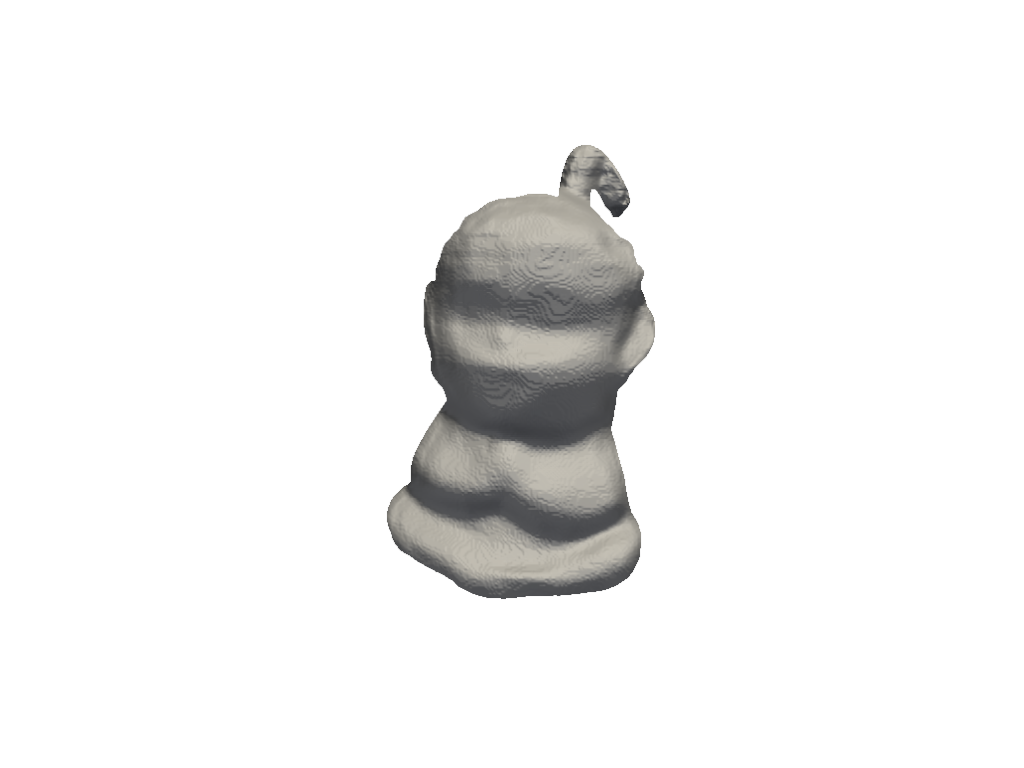}
         &\includegraphics[width = \imagesize\linewidth, trim = 355pt 170pt 340pt 140pt, clip]{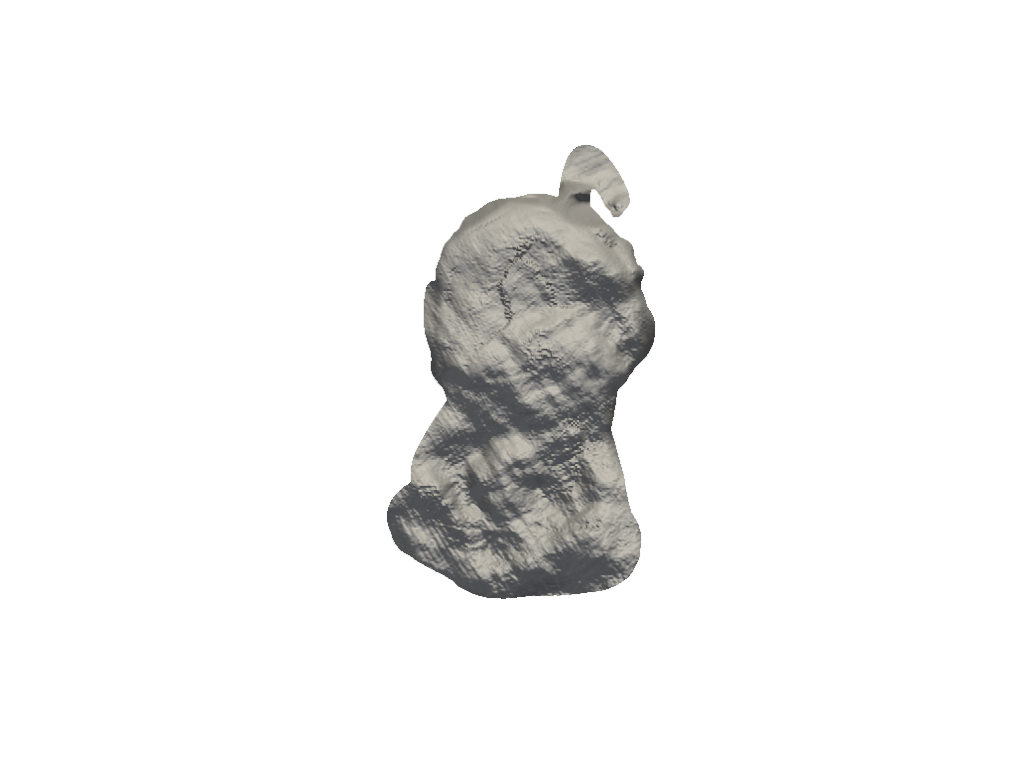}
         &\includegraphics[width = \imagesize\linewidth, trim = 355pt 170pt 340pt 140pt, clip]{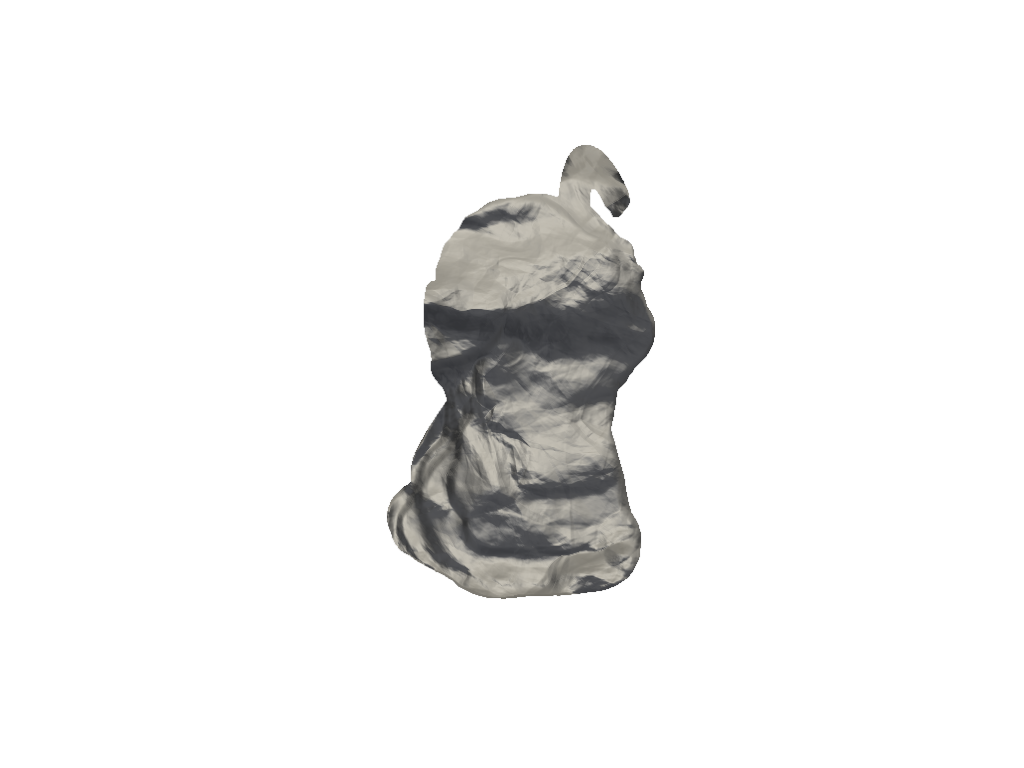}
         &\includegraphics[width = \imagesize\linewidth, trim = 355pt 170pt 340pt 140pt, clip]{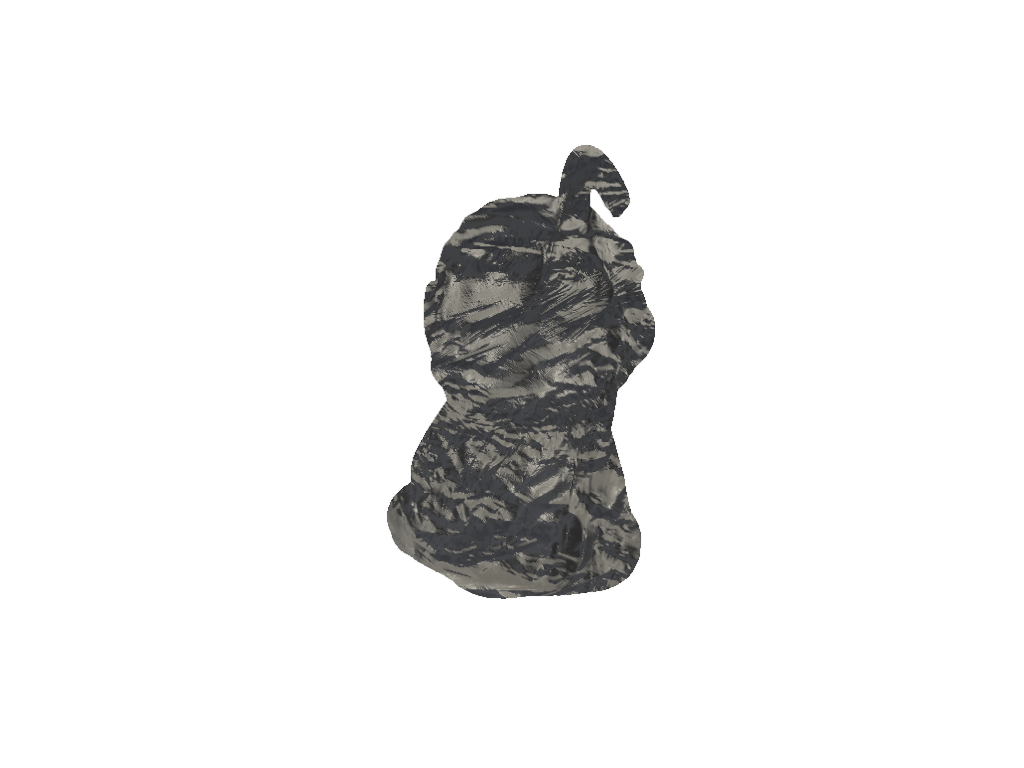}
         &\includegraphics[width = 0.084\linewidth, trim = 0pt 0pt 0pt 0pt, clip]{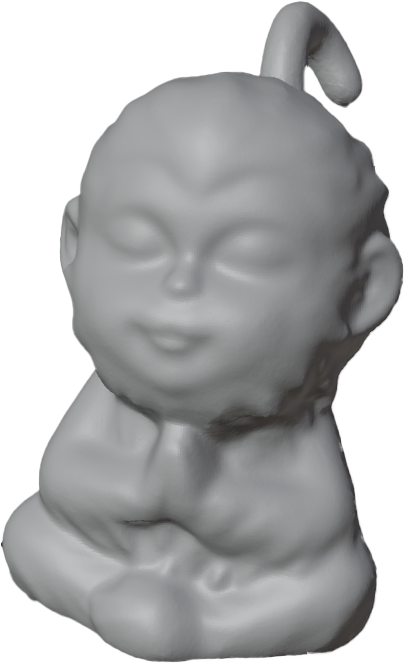}
        \\
         \includegraphics[width = 0.115\linewidth, trim = 253pt 110pt 300pt 110pt, clip]{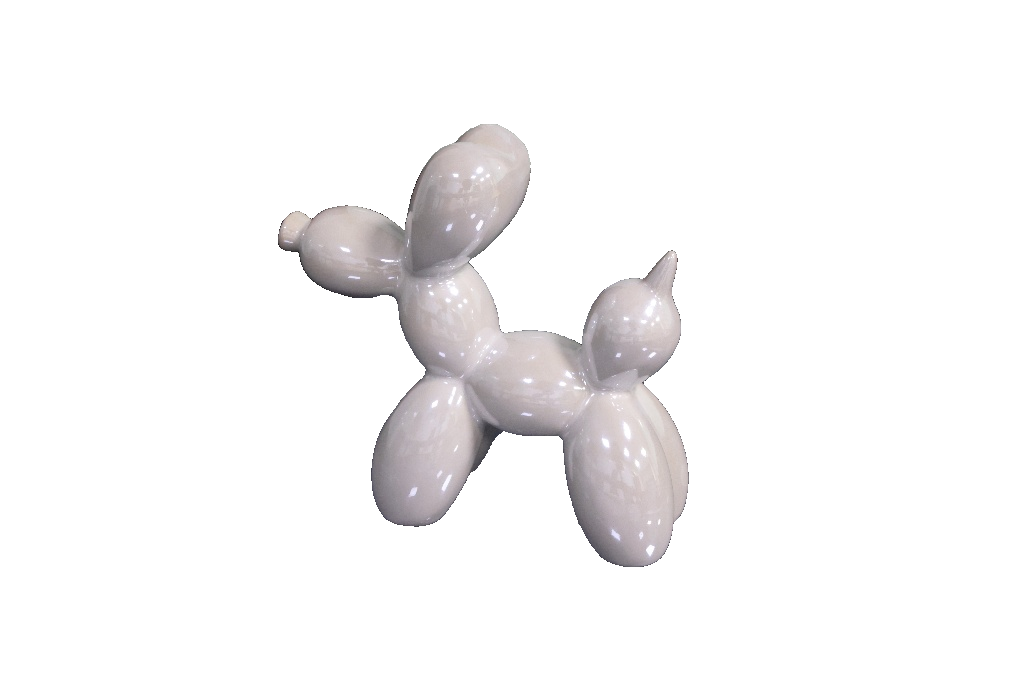}
         &\includegraphics[width = 0.096\linewidth, trim =  0pt 0pt 0pt 0pt, clip]{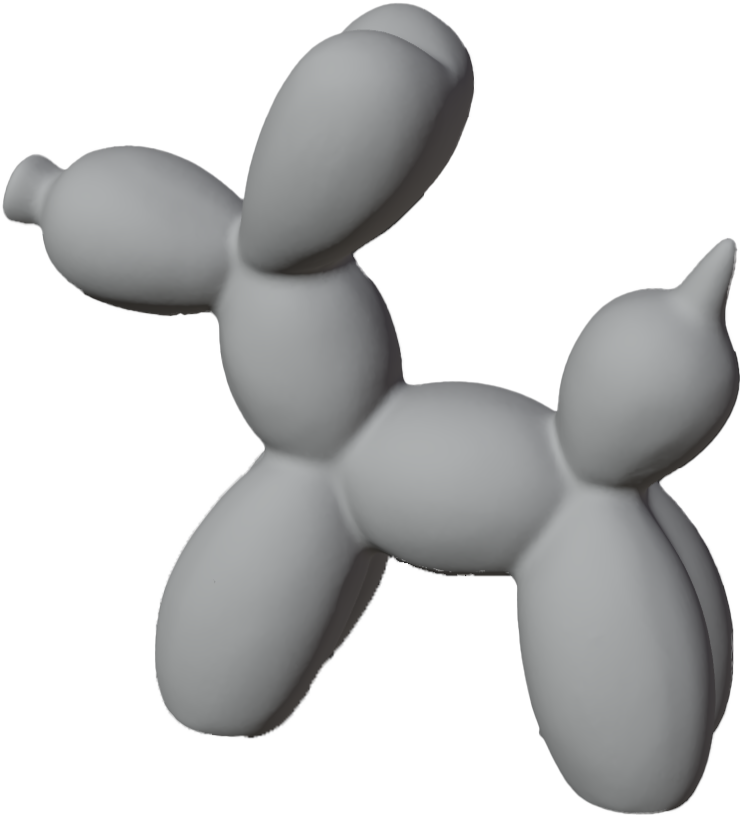}
         &\includegraphics[width = 0.096\linewidth, trim =  0pt 0pt 0pt 0pt, clip]{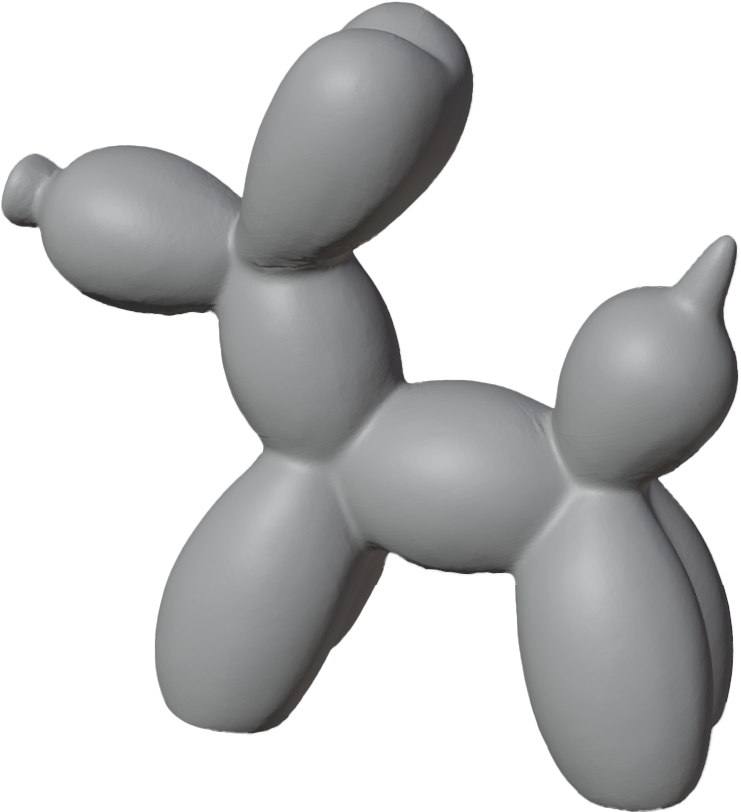}
         &\includegraphics[width = \imagesize\linewidth, trim =  253pt 155pt 300pt 150pt, clip]{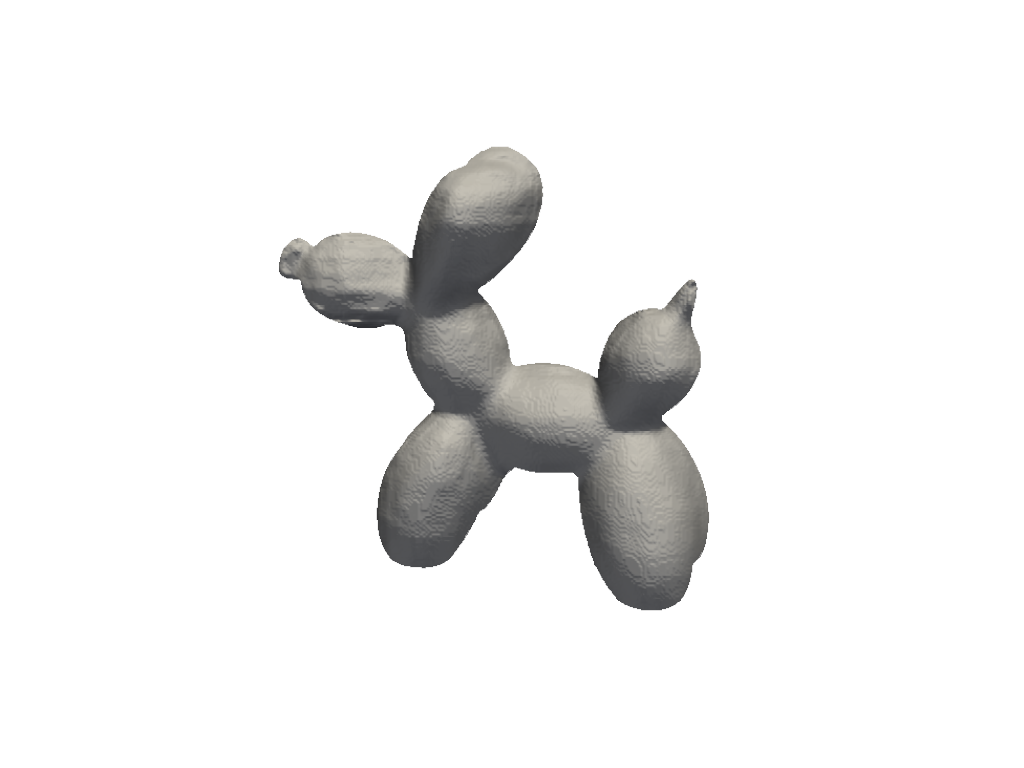}
         &\includegraphics[width = \imagesize\linewidth, trim =  253pt 155pt 300pt 150pt, clip]{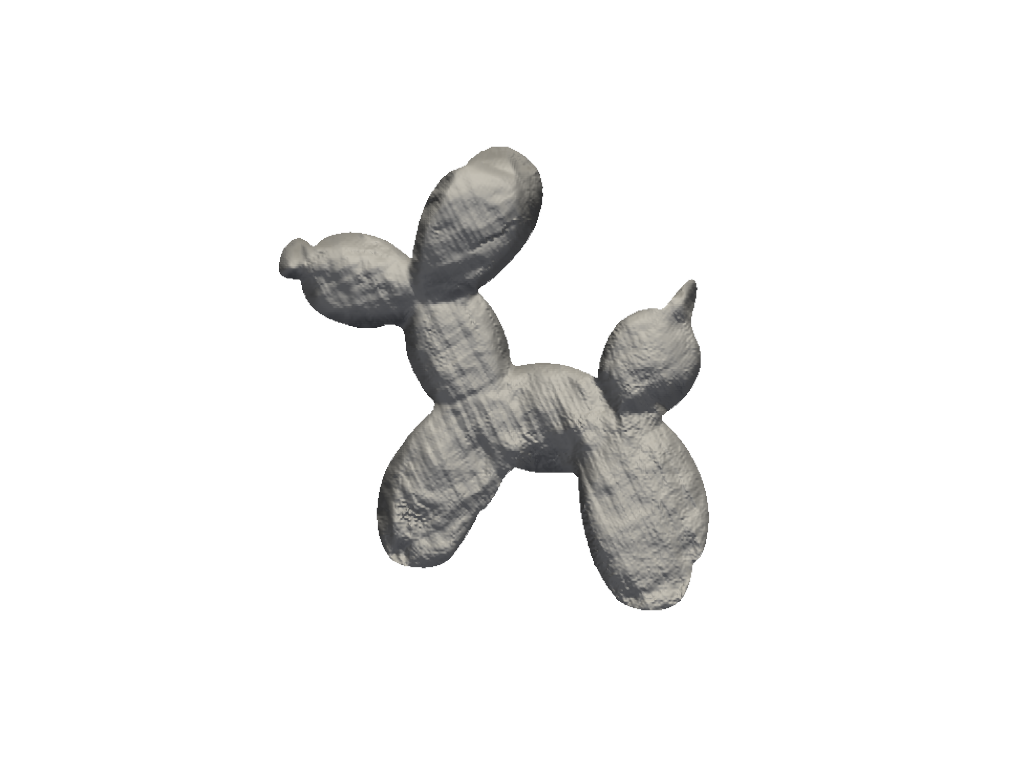}
         &\includegraphics[width = \imagesize\linewidth, trim =  253pt 155pt 300pt 150pt, clip]{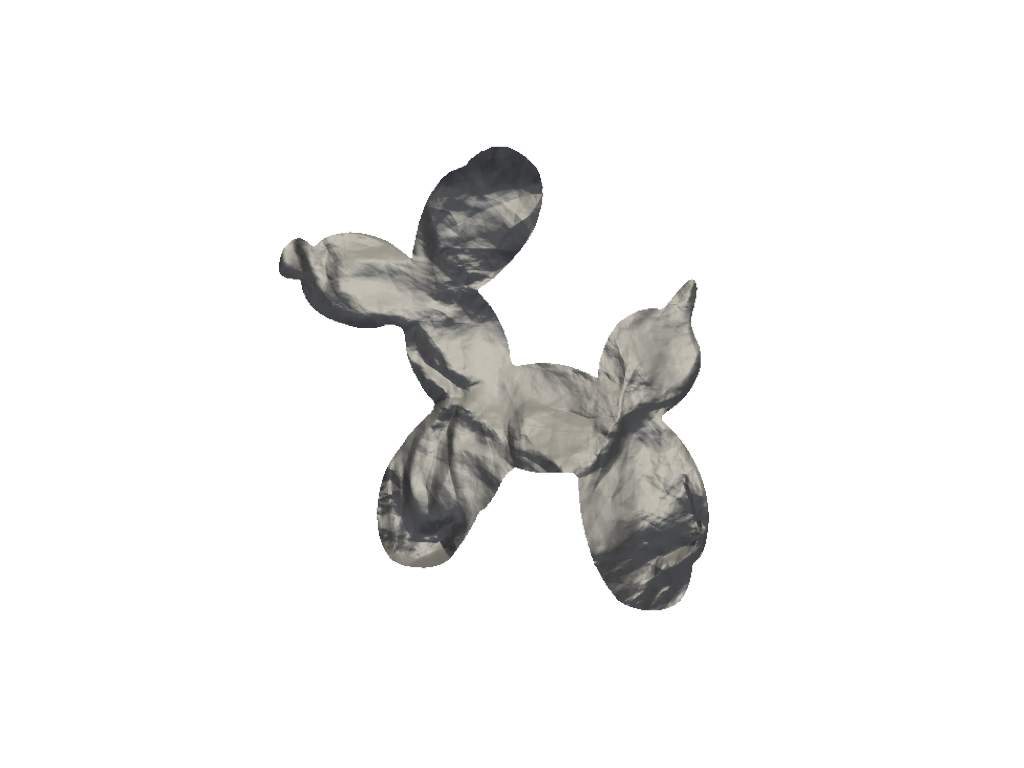}
         &\includegraphics[width = \imagesize\linewidth, trim =  253pt 155pt 300pt 150pt, clip]{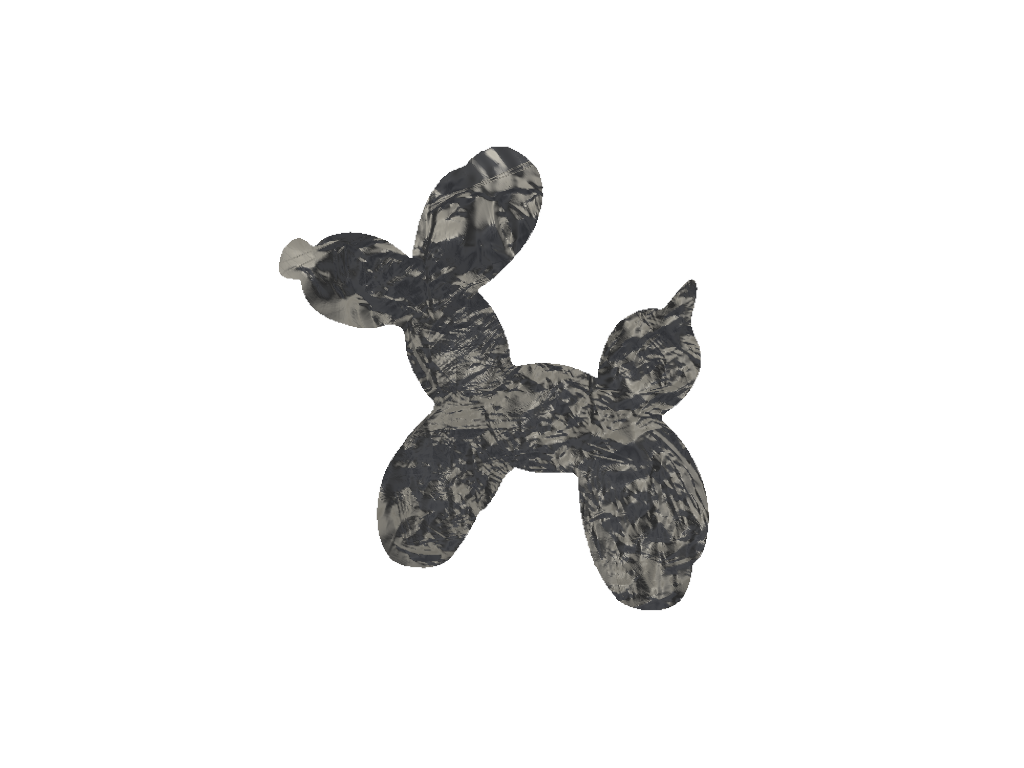}
         &\includegraphics[width = 0.096\linewidth, trim =  0pt 0pt 0pt 0pt, clip]{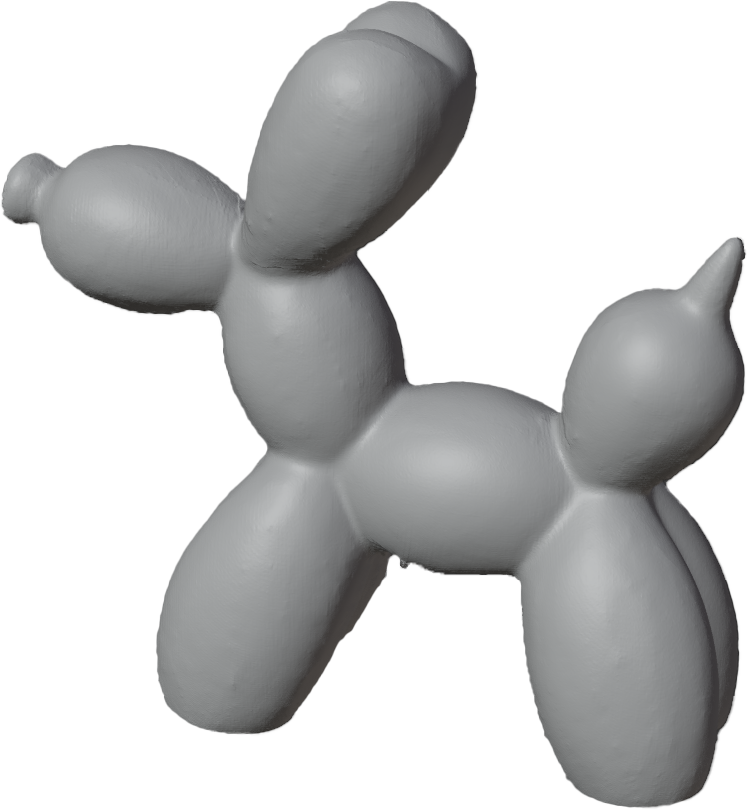}
    \end{tabular}}
	\caption{Qualitative evaluation of shape recovery on {\sc Monkey} and {\sc Dog} objects of the \rgreal dataset. }
	\label{fig:shape_est}
\end{figure*}

\begin{table}
	\caption{Ablation study on different loss terms.}
	\resizebox{\linewidth}{!}{
	\begin{tabular}{@{}lcccc@{}}
		\toprule
	\multirow{2}{*}{Method}	& \multicolumn{2}{c}{Shape estimation} & \multicolumn{2}{c}{Poss estimation} \\
		& CD~$\downarrow$ & F1-score~$\uparrow$ &RPEt~$\downarrow$ &RPEr($^\circ$)~$\downarrow$ \\
		\midrule
        Ours w/o $\mathcal{L}_{normal} $ & 0.691 & 0.769 &  0.139 &  0.542\\
		Ours w/o $\mathcal{L}_{ni} $ & 0.163 & 0.988 & 0.082 & 0.280\\
		Ours w/o $\mathcal{L}_{c} $ & 0.126 & 0.998 & 0.052 & 0.210 \\
		Ours &  \textbf{0.115} & \textbf{0.999} & \textbf{0.037} & \textbf{0.141}\\
		\bottomrule
	\end{tabular}
	}
    \label{tab:ablation}
    \vspace{-0.5em}
\end{table}

\subsection{Ablation study}
We conduct an ablation study to test the effectiveness of different loss terms, taking {\sc Pot2} in \dmv as an example. As shown in \Tref{tab:ablation}, 
$\loss_{normal}$ contributes most to the shape and pose recovery, demonstrating the necessity of supervising surface normal for pose-free 3D reconstruction. 
Without $\loss_{ni}$, the prior of integrated depth is missing, leading to an error increase in shape and pose estimation.
$\loss_{c}$ based on multi-view surface normal consistency also helps to improve the accuracy. Combining these loss terms, our method get accurate shape and pose estimation.

\section{Conclusion}
We introduce PMNI, the first method that recovers both shape and pose solely from multi-view normal maps. Due to the scarcity of features in reflective and textureless objects in the RGB domain, existing joint optimization-based methods struggle with pose and shape recovery. In contrast to RGB images, \ours utilizes surface normals as input, which are robust to reflective and textureless surfaces. By incorporating depth from normal integration as a prior and leveraging multi-view geometric consistency, we jointly optimize shape and camera poses using a neural SDF network. We hope our method will contribute to detailed 3D reconstruction in casual capture settings.

\section*{Acknowledgment} This work was supported by Hebei Natural Science Foundation Project No. F2024502017, Beijing-Tianjin-Hebei Basic Research Funding Program No. 242Q0101Z, National Natural Science Foundation of China (Grant No. 62472044, U24B20155, 62225601, U23B2052), Beijing Natural Science Foundation (Grant No. QY24202). We thank openbayes.com for the support of computational resources.

{
    \small
    \bibliographystyle{ieeenat_fullname}
    \bibliography{main}
}

\end{document}